\pdfoutput=1

\documentclass[11pt]{article}

\usepackage{acl}

\usepackage{times}
\usepackage{latexsym}

\usepackage[T1]{fontenc}

\usepackage[utf8]{inputenc}

\usepackage{microtype}

\usepackage{inconsolata}

\usepackage{booktabs}
\usepackage{graphicx}
\usepackage{multirow}
\usepackage{bbm}
\usepackage{makecell}
\usepackage{xcolor}
\graphicspath{{figures/}}
\usepackage{lipsum}

\usepackage{bbm}
\usepackage{amsmath}
\usepackage{float}
\usepackage{mathtools}
\usepackage{subcaption}
\usepackage{algorithm, algpseudocode}
\usepackage{newunicodechar}
\usepackage{footnotehyper, tablefootnote}
\usepackage{svg}
\usepackage{amsfonts}
\usepackage{scalerel}
\usepackage[separate-uncertainty,group-minimum-digits=4]{siunitx}

\def\divorced{\scalerel*{\includegraphics{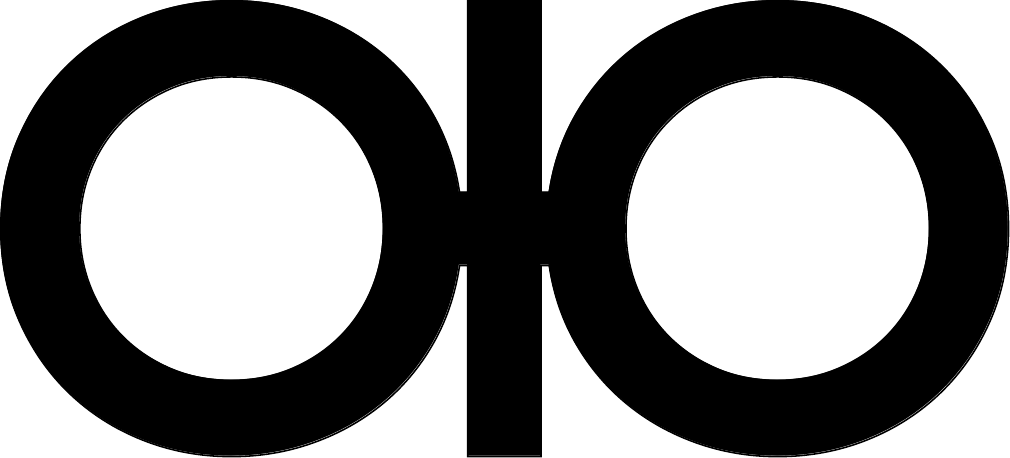}}{X}}
\def\unmarried{\scalerel*{\includegraphics{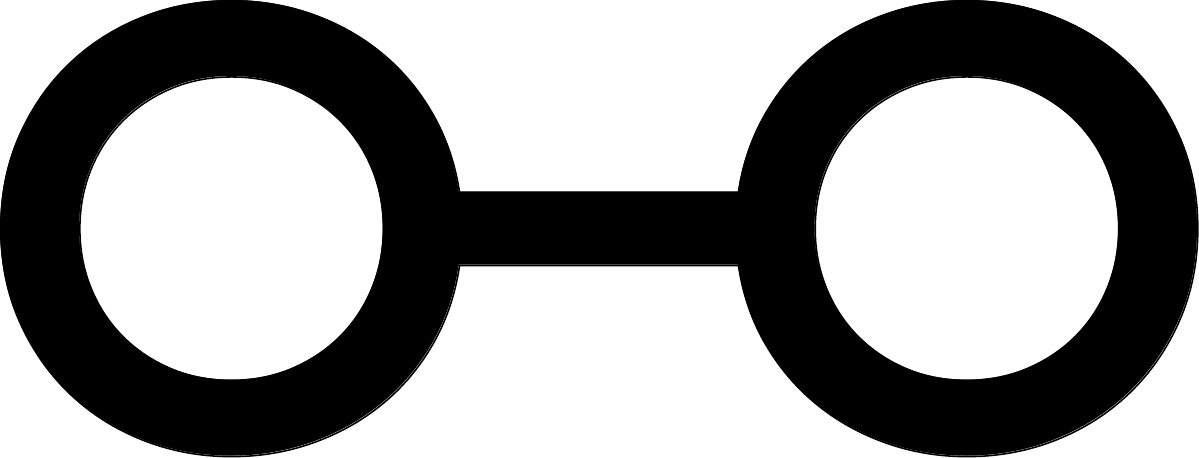}}{X}}

\usepackage{booktabs}
\usepackage{cleveref}

\usepackage{marvosym}

\usepackage{enumitem} 

\title{Déjà Vu? Decoding Repeated Reading from Eye Movements}

\author{Yoav Meiri$^1$, Omer Shubi$^1$, Cfir Avraham Hadar$^1$\\ {\bf Ariel Kreisberg Nitzav$^1$ Yevgeni Berzak$^{1,2}$} \\
 $^1$Faculty of Data and Decision Sciences, \\
 Technion - Israel Institute of Technology, Haifa, Israel \\
 $^2$Department of Brain and Cognitive Sciences, \\
 Massachusetts Institute of Technology, Cambridge, USA \\
 \texttt{\{meiri.yoav,shubi,kfir-hadar,ariel.kr\}@campus.technion.ac.il} \\
\texttt{berzak@technion.ac.il} \\
}

\begin{document}
\maketitle
\begin{abstract}
Be it your favorite novel, a newswire article, a cooking recipe or an academic paper -- in many daily situations we read the same text more than once. In this work, we ask whether it is possible to automatically determine whether the reader has previously encountered a text based on their eye movement patterns. We introduce two variants of this task and address them with considerable success using both feature-based and neural models. We further introduce a general strategy for enhancing these models with machine generated simulations of eye movements from a cognitive model. Finally, we present an analysis of model performance which on the one hand yields insights on the information used by the models, and on the other hand leverages predictive modeling as an analytic tool for better characterization of the role of memory in repeated reading. Our work advances the understanding of the extent and manner in which eye movements in reading capture memory effects from prior text exposure, and paves the way for future applications that involve predictive modeling of repeated reading.\footnote{Code is available anonymously \href{https://anonymous.4open.science/r/deja-vu}{here}. 
}
\end{abstract}

\section{Introduction}

Reading is a widely practiced skill that occupies many hours of our daily lives. During these hours, there are various ways in which we interact with texts. 
While reading is often thought of as an interaction with new linguistic material, in many daily scenarios we read texts more than once. This can happen because we might want to understand or recall the text better, re-examine specific parts of interest, or simply because we enjoyed reading the text for the first time. 


The importance of studying repeated reading for understanding human language processing has long been recognized in psychology and psycholinguistics. In these areas of study, it was shown that when reading a text for a second time, the way our eyes move over the text and the extent to which the eye movements depend on the linguistic characteristics of the text tend to differ compared to the first reading. In essence, eye movements in repeated reading reflect reading facilitation: for example, readers tend to read faster and skip more words compared to the first reading. Although the precise differences can depend on the experimental setup, and some are debated, the presence of facilitation effects comes as no surprise, as when encountering a text for the second time readers already have knowledge of its content, and can more easily foresee what comes next at any given time.

Despite the advances in the study of eye movements in repeated reading, prior work has been limited to \emph{descriptive} analyses of overall effects, averaged across texts and participants. Consequently, it is currently unknown how much information can be extracted regarding the type of interaction of a specific reader with a specific text. Addressing this question is important both for improving the scientific understanding of the extent and manner in which eye movements reflect the reader's memory of the text, and for building the foundations for practical applications in areas such as e-learning and educational settings more broadly, where it can be beneficial to infer whether the reader has already encountered the text. 

In this work, we tackle this challenge using a \emph{predictive} modeling approach for determining the interaction of a single reader with a specific text from their eye movements. We pose the following question: is it possible to decode whether the reader is reading a text for the first or the second time from their eye movements over the text? Addressing this question is made possible by OneStop Eye Movements \citep{onestop2025preprint}, the first publicly available dataset that contains eye movement recordings of both first and repeated reading.

We operationalize this question via two prediction tasks. In the first task, given two eye movement samples from the same participant over the same text, the goal is to determine which is first reading and which is repeated reading. In the second, more general, and more challenging variant, the task is to determine whether a single eye movement sample is a first or repeated reading. We address these tasks with a feature-based approach with features that build on the psycholinguistic literature on repeated reading, and with multimodal neural models that combine eye movements with text. We further introduce a strategy for improving predictive accuracy by augmenting models with synthetic eye movement trajectories from a cognitive model of eye movements in reading. 

The contributions of this work are the following:
\begin{itemize}[leftmargin=*]
 \item \textbf{Tasks}: We introduce a new prediction task for the interaction of the reader with the text from their eye movements - automatically determine whether the reader encountered the text previously. We address this task in two variants of decreasing difficulty: (i) a single eye movement sample; (ii) a pair of first and second reading samples from the same participant.
 \item \textbf{Modeling}: We experiment with two types of predictive approaches: (i) feature-based models; (ii) neural multimodal language models. We further introduce a strategy for integrating into the prediction pipeline synthetic data for first reading.
 \item \textbf{Analyses}: We present analyses of model performance as a function of article location in the experiment and the amount of intervening material between readings. These analyses provide insights on the information used by models and on the role of memory in repeated reading effects.
\end{itemize}

\section{Related Work}
\label{sec:related-work}

\begin{figure}
    \centering
    \includegraphics[width=0.85\linewidth]{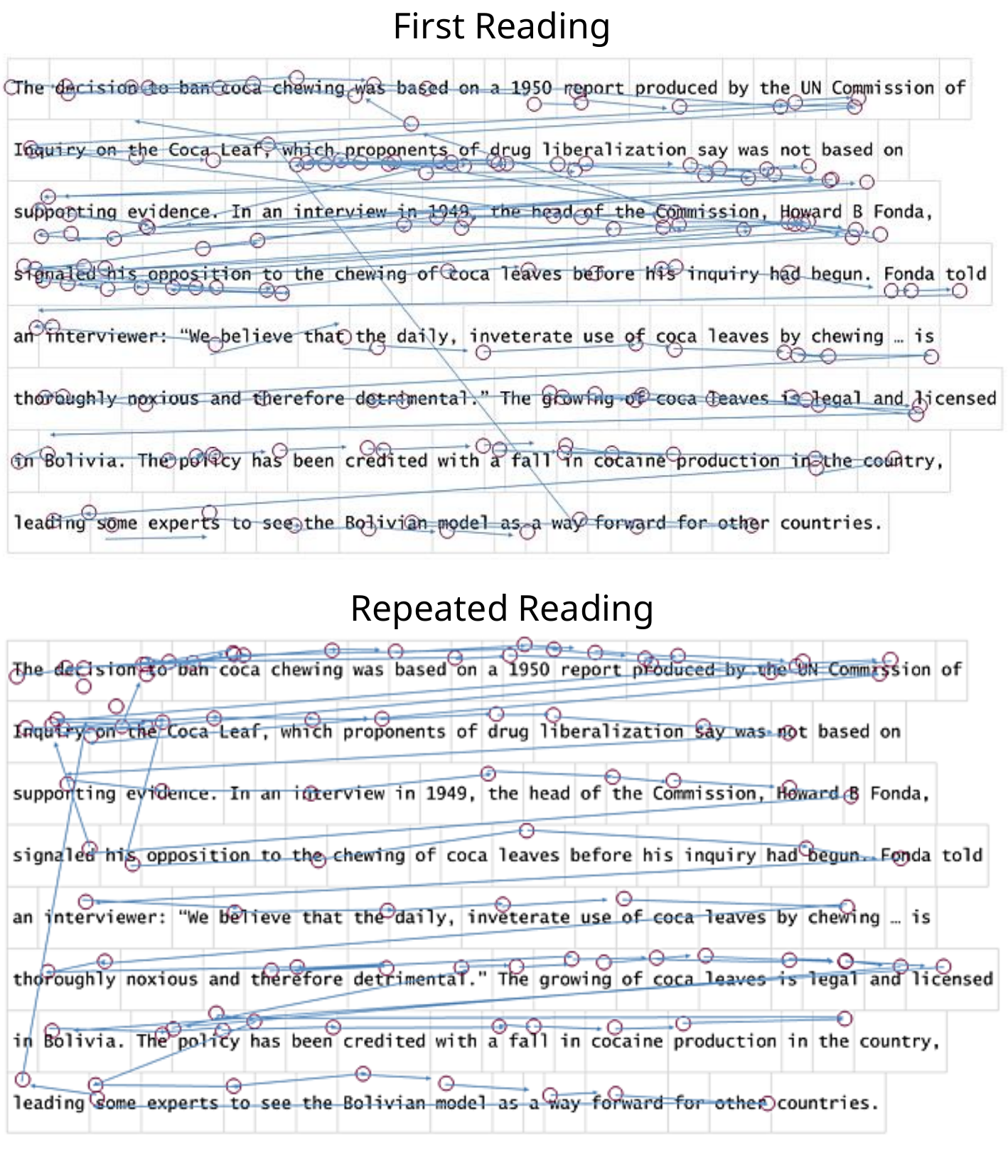}
    \caption{Examples of eye movements over a single passage; top: first reading, bottom: repeated reading. Circles represent fixations, and lines represent saccades.}
    \label{fig:trial-example}
\end{figure}

When we read, the eye movement trajectory, or \emph{scanpath} over the text is divided into \emph{fixations}, prolonged periods of time during which the gaze location is relatively fixed, and \emph{saccades}, fast transitions between fixations \cite{rayner1998eye,schotter2025beginner}. Prior work in psycholinguistics has consistently demonstrated that this trajectory differs in repeated reading compared to first reading, with large \emph{facilitation} effects marked by shorter text reading times, fewer fixations, shorter fixation durations, longer saccades and fewer regressions (backward saccades) \citep{hyona1990repeated,raney1995freq,schnitzer2006,meiri2024deja}. Several studies have also examined the interaction between repeated reading and the effect of linguistic word properties such as word length, frequency and surprisal on reading times, mostly finding less sensitivity to word properties in repeated reading \citep{raney1995freq,foster2013repeated,zawoyski2015repeated,meiri2024deja}. In line with these studies, \citet{hyona1990repeated} further demonstrated a reduction in the sensitivity of eye movements to the introduction of new topics in rereading. All the above studies examined individual features aggregated across participants and texts, and it is currently unknown whether first and repeated reading can be effectively distinguished using predictive modeling at the level of an individual participant and text.


In machine learning and NLP, a nascent line of work focuses on decoding the properties of the reader and their interaction with the text, from eye movements in reading. These include, among others, decoding of linguistic knowledge \citep{berzak_predicting_2017,berzak_assessing_2018,skerath2023native}, reading comprehension \cite{ahn_towards_2020,reich_inferring_2022,meziere2023using,Shubi2024Finegrained}, subjective text difficulty \cite{reich_inferring_2022} and the reader's goals \citep{hollenstein2023zuco,shubi2024decoding}. The current study falls broadly within this area, but introduces and addresses a new task of decoding repeated reading. 
Following \citet{sood2020improving}, our work leverages the output of \mbox{E-Z} Reader, a computational cognitive model for automatic generation of reading scanpath trajectories \cite{reichle1998toward,reichle2003ez,reichle2009using,veldre2023understanding}.

\section{Problem Formulation}
\label{sec:problem}

We ask whether it is possible to accurately distinguish between first and second readings, 
from an eye movement recording of a single participant over a single textual item.
We assume a setup in which a participant \(S\) reads a textual item \(W\), optionally reads \(k\) other items \(\{W'\}^k\), and then reads \(W\) again. The parameter \(k\) can range from 0 for consecutive repeated reading to any $k>0$ for non-consecutive repeated reading. 
Hence, each reading \(r \in \{1,2\}\) (first or repeated) of \(W\) produces a distinct eye movement recording \(E_S^{W,r}\). We define a \emph{decoding task} where the goal is to distinguish between eye movements of a single participant over a single text in first reading \((r=1)\) and repeated reading \((r=2)\). The task has two variants as described below.

\paragraph{Single Trial Task}
The input is an eye movement recording \(E_S^{W,r}\) for text $W$.
The output \(\hat{r} \in \{1,2\}\) corresponds to whether the eye movements \(E\) are from a first or a repeated reading of \(W\). Formally,
\[
(W,\; E_S^{W,r}) \;\longrightarrow\; \hat{r}
\]

\paragraph{Paired Trials Task}
Here the input consists of two eye movement recordings \(E_S^{W,r}\) and \(E_S^{W,r'}\) of the same participant \(S\) in an unknown presentation order and the same text \(W\). 
The output is \(\quad  \hat{\left(r,r'\right)} \in \{\left(1,2\right),\left(2,1\right)\}\), i.e., which recording corresponds to the first reading of $W$, and which to the second. Formally,
\[
(W,\; E_S^{W,r},\; E_S^{W,r'}) \;\longrightarrow\; \hat{\left(r,r'\right)} 
\]

\section{Data}
\label{sec:data}

We use OneStop Eye Movements \cite{onestop2025preprint} an eyetracking dataset collected with an Eyelink 1000 Plus eyetracker, where native (L1) speakers of English read Guardian newswire articles. The textual materials are taken from the OneStopQA dataset \citep{berzak_starc_2020}. OneStop Eye Movements includes 180 participants who read for comprehension, each reading a 10-article batch in a randomized order, where each article contains between 4 and 7 paragraphs. Participants read each paragraph on a single page without the ability to return to previous paragraphs. 

\begin{figure}
    \centering
    \includegraphics[width=0.65\linewidth]{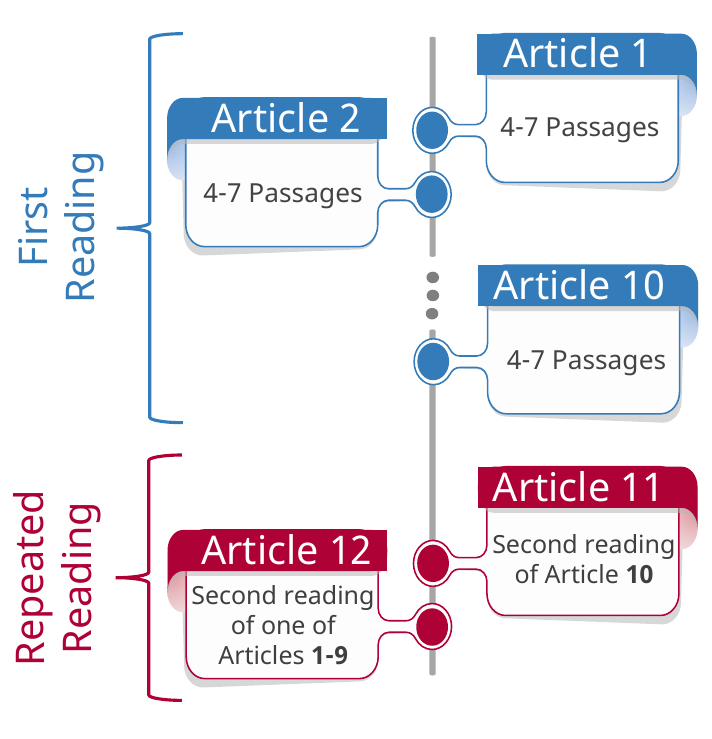}
    \caption{First and repeated reading for one participant. After reading a 10-article batch in a random order of articles, there is a consecutive repeated reading of the last article in position 10, and then a non-consecutive repeated reading of one of the articles in positions 1-9.}
    \label{fig:exp-schema}
\end{figure}

After reading a 10-article batch, participants read two articles for a second time. In repeated reading, the paragraphs are identical to the first reading, while the questions are different. The article in position 11 is a consecutive second presentation of the article in position 10. The article in position 12 is a non-consecutive second presentation of an article in one of the positions 1-9. Thus, half of the repeated reading data captures immediate consecutive rereading of the same article, and the other half is rereading with intervening reading material, ranging from 2 to 10 articles. \Cref{fig:exp-schema} presents the experimental design schematically.

Overall, there are 360 second presentations of articles, 180 in consecutive rereading in position 11 and 180 in a non-consecutive rereading in position 12. The first reading of position 12 articles occurs 36 times in position 1 and 18 times in each of the positions 2-9. The 360 repeated article readings correspond to 1,944 paragraph trials with a total of 105,540-word tokens over which eye movements were collected, split equally between positions 11 and 12. \Cref{fig:trial-example} shows example trials for first reading and repeated reading. \Cref{sec:app-data} presents further details on the data. 

\section{Modeling}

We experiment with both feature-based and neural language modeling approaches for representing eye movements and their interaction with the text. 
For the single trial variant, we present methods that directly predict first versus repeated reading. We also propose a general approach for leveraging synthetic scanpaths as an additional first reading reference, which enables using the same input representations as in the paired trials task.






\subsection{Single Trial Modeling}

\subsubsection*{Feature-Based Model}

For feature-based modeling, we use XGBoost tree-boosting models \cite{chen2016xgboost} with the following global eye movement features, resulting in an input feature vector $e_S^{\text{global}} \in \mathbb{R}^{d_{\text{global}}}$ where $d_{global}=35$. The features are motivated by the psycholinguistic literature in general, and work on differences between first and repeated reading in particular. See \Cref{app:trial_level_features} for further details on the trial-level features and training procedure.

\begin{itemize}[leftmargin=*]
 
 \item \textbf{Standard Eye Movement Measures} 8 standard eye movement measures from the psycholinguistic literature, including per word averages of Total Fixation Duration, First Fixation Duration, Gaze Duration, number of Fixations, skip rate and regression rate. See \Cref{app:word_fix_level_features} for additional features and definitions. These features were previously used for prediction tasks from eye movements \citep[e.g.][]{meziere2023using} and were shown to differ between first and repeated reading (see \Cref{sec:related-work}). 
      
 \item \textbf{Word Property Coefficients} 20 features that measure the responsiveness of reading measures to linguistic word properties: frequency, surprisal and length. Building on \citet{berzak_assessing_2018}, the features are coefficients from linear models that predict the participant's speed-normalized eye movement measures from these three word properties. This feature-set is motivated by prior work that has demonstrated that the responsiveness of eye movements to linguistic word properties varies across reading scenarios and readers \citep[e.g.][]{reichle_eye_2010,berzak2023traces,shubi2023}. In repeated reading, this responsiveness is weaker compared to first reading 
 \cite{meiri2024deja}. See \Cref{app:response_to_ling} for further details on the models.
 
 \item \textbf{Saccade Network Measures}\label{sec:feature_based_methods} 
 Following \citet{zhu2015exploratory}, we define a directed graph that encodes the scanpath of eye movements over the paragraph $G=\{V,T\}$ such that $V$ is the set of words in the paragraph, and for all $u,v\in V$:
 $$T=\{\left(u,v\right): \text{there is a transition from }u \text{ to }v\}$$ We extract 7 features which capture connectivity, centrality and clustering measures of this graph.
 Additional details and definitions of the measures, along with network visualization examples are provided in \Cref{app:network_measures}.

\end{itemize}

\subsubsection*{Neural Models}

We use two variants of the \textbf{RoBERTEye} multimodal language model  \citep{Shubi2024Finegrained}, which is a state-of-the-art approach in predictive modeling using eye movements in reading. This model was previously applied to the prediction of reading comprehension \citep{Shubi2024Finegrained} and reading goals \citep{shubi2024decoding} from eye movements, outperforming in most cases prior models from the literature.
RoBERTEye extends the RoBERTa model \citep{liu_roberta_2019} by incorporating eye movement information. It does so by projecting an input eye movement feature vector for each word/fixation into the embedding space of the language model and then concatenating these projections with the word embedding sequence. 

The model has two variants, with \textbf{word-level} and \textbf{fixation-level} eye movement representations. In \textbf{RoBERTEye-Words} 
the eye movements input consists of \( \left(e_S^{\text{word}_j} \right)_{j=1}^{N_{\text{words}}} \) where each \(e_S^{\text{word}_j} \in \mathbb{R}^{d_{\text{word}}} \) is an eye movement feature vector for the word $j$, with $d_{word}=13$ features. 

In \textbf{RoBERTEye-Fixations}, both fixation-level and word-level features are used. Each fixation $i$ on word $j$ has a fixation vector \( e_S^{\text{fix}_{i,j}} \in \mathbb{R}^{d_{\text{fix}}} \) with $d_{fix}=6$ features of the fixation. 
This vector is concatenated with the word-level feature vector \( e_{S}^{\text{word}_j} \):
\[
\left( e_S^{\text{fix}_{i,j}} \oplus e_S^{\text{word}_{j}} \right)_{i=1}^{N_{\text{fixations}}}
\]
where \( \oplus \) denotes the concatenation operation along the feature dimension. To help the model distinguish between eye movement and textual information, two special token vectors are added, one to all the text embeddings, and the other to all the projected eye movement embeddings. 
Complete lists for both fixation-level and word-level features are provided in \Cref{app:word_fix_level_features}.

\subsection{Single Trial Modeling with Synthetic Scanpath References}


We introduce a new method, where in addition to the human eye movement data, the model input further includes a \emph{synthetic scanpath} reference \(E_M^{W,1}\) of eye movements  \(E\) generated for each text $W$ from an external model \(M\) for scanpath generation. 
As all existing computational models for scanpath generation assume a first reading, in this work we focus on the generation of first reading reference scanpaths. In essence, this reference provides an external source of information on how a typical first reading should looke like. This addition of the reference yields a task whose input structure resembles the paired trials task, only that one of the eye movement inputs is now machine generated:
\[
(W,\; E_M^{W,1} \; E_S^{W,r}) \;\longrightarrow\; \hat{r}.
\]
We then obtain the following three types of representations of eye movements:
\paragraph{Feature-based representations} The representation is a concatenation of the synthetic features, and their difference from the human features. Formally,
\[
 e_S^{\text{global}} \oplus (e_M^{\text{global}} -e_S^{\text{global}})
\]
\paragraph{Word-level representations} 
For each word, we concatenate its machine generated features with their difference from the human features. Formally,
\[
 \left( e_S^{\text{word}_j} \oplus (e_M^{\text{word}_j} -e_S^{\text{word}_j})\right)_{j=1}^{N_{words}}
\]

\paragraph{Fixation-level representations} Unlike the global and word-level representations, fixation-level features are not aligned. We therefore construct the input as follows
\[
\left( e_S^{\text{fix}_{i,j}} \oplus e_S^{\text{word}_{j}} \right)_{i=1}^{N_{\text{fixations}}} \mathbin\Vert \left( e_M^{\text{fix}_{i,j}} \oplus e_M^{\text{word}_{j}} \right)_{i=1}^{\tilde{N}_{\text{fixations}}}
\]
where \(\mathbin\Vert\) denotes concatenation along the sequence dimension, and \(\tilde{N}_{\text{fixations}}\) is the length of the scanpath generated by \(M\).
To help RoBERTEye distinguish between human and synthetic scanpath features, in addition to the word and human eye movement special tokens, we introduce a third token that marks machine generated scanpaths.

\paragraph{\mbox{E-Z} Reader Scanpaths} The synthetic scanpaths are generated using \mbox{E-Z} Reader \cite{reichle1998toward,reichle2003ez,reichle2009using,veldre2023understanding}, a prominent computational cognitive model for eye movements generation. The full details of the generation process and the adaptations made to the original model are discussed in \Cref{app:synthesis_details}. As we expect the effectiveness of the augmentation approach to depend on the quality of the generated scanpaths, we perform an evaluation of \mbox{E-Z} Reader outputs in the context of our task. 
To this end, we compare \mbox{E-Z} Reader outputs with human eye movements in both first and repeated reading. Our analysis examines the overall similarity of the scanpaths, which we expect to be greater in first reading, as well as the direction of the deviations.
A necessary condition for \mbox{E-Z} Reader outputs to be effective as approximations of first reading behavior, is that on average, they should be more similar to human first reading than to human repeated reading.  

\Cref{tab:gen-quality} suggests that this indeed tends to be the case. It presents four measures for which robust differences between first and repeated reading were previously observed \citep{meiri2024deja}. Using mixed-effects models with text-level bootstrapping (see \Cref{app:sec:synth-data-analysis}), we compared the absolute differences of  \mbox{E-Z} Reader from first and repeated reading.
For Fixation Count and Skip Rate, \mbox{E-Z} Reader is significantly closer to first reading ($p<0.001$), whereas Regression Rate is significantly closer to repeated reading ($p<0.001$).
Although mean Total Fixation Duration (TF) does not show a significant difference ($p\approx0.53$), the direction of the difference remains useful for classification. Overall, these findings support the viability of \mbox{E-Z} Reader as an approximation of human first reading scanpath trajectories.

\begin{table}[ht]
\small
\resizebox{\columnwidth}{!}{%
    \setlength{\tabcolsep}{2pt}%
    \centering
    \begin{tabular}{
        l
        c
        c
    }
        \toprule
        \makecell[l]{\textbf{Measure}} & \makecell{\textbf{\textcolor[HTML]{347BB9}{First Reading}}} & \makecell{\textbf{\textcolor[HTML]{AE0135}{Repeated Reading}}} \\
        \midrule
        \makecell[l]{Fixation Count} & $0.03_{\pm0.01}$ & $-0.31_{\pm0.01}$ \\
        \addlinespace
        \makecell[l]{Mean TF (ms)} & $27_{\pm2.3}$  & $-29_{\pm1.7}$ \\
        \addlinespace
        \makecell[l]{Regression Rate} & $0.2_{\pm0.003}$ & $0.1_{\pm0.002}$ \\
        \addlinespace
        \makecell[l]{Skip Rate} & $0.2_{\pm0.003}$ & $0.3_{\pm0.003}$ \\
        \bottomrule
    \end{tabular}%
}
\caption{Trial-level mean differences between human and E-Z Reader-generated measures for four standard eye movement metrics, with 95\% confidence intervals.}

\label{tab:gen-quality}
\end{table}

\begin{figure*}
    \centering
    \includegraphics[width=1\linewidth]{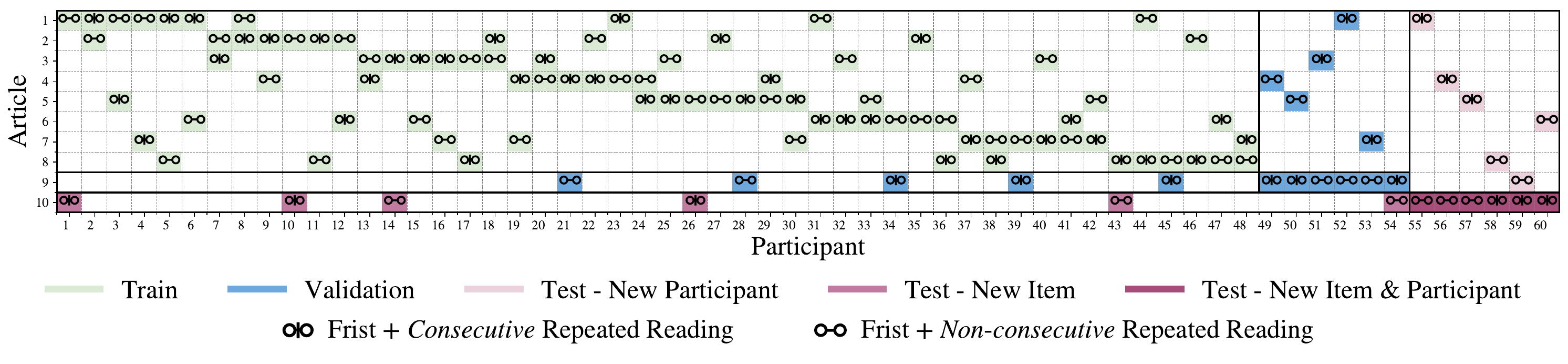}
    \caption{Visualization of a 10-article, 60-participant data split, divided into train, validation, and three test regimes.  
    Each non-empty cell represents a participant-article pair, comprising the first and repeated readings of an article by the same participant. '{\scriptsize \protect\divorced}' denotes consecutive repeated reading and '{\scriptsize \protect\unmarried}' denotes non-consecutive repeated reading (i.e. with intervening articles between the first and second readings).}
    \label{fig:CV}
\end{figure*}

\mbox{E-Z} Reader is a probabilistic model that samples scanpaths for a given text. We generate 1000 synthetic scanpaths for each paragraph, and then augment the human eye movement data with reference features derived from these first reading simulated scanpaths. To obtain global and word-level representations, we first average measures across all the generated scanpaths. 
The averaging aims to enhance the robustness of the representations by reducing noise inherent to a single scanpath. 
For the fixation-level representation, averaging scanpaths is not applicable, and we therefore follow  \citet{meziere2024scanpath} in selecting a \emph{prototype scanpath} that minimizes the mean scanpath distance to all other scanpaths, using the Scasim scanpath similarity metric \cite{von2011scanpath}. 
\subsection{Paired Trials Modeling}

In the paired trials task, we use the same feature representation as in the single trial task with machine generated scanpaths described above, only that now both eye movement samples are from a human participant, and the third special token marks the second human input. Further, differently from the single-augmented setting, the order of the two inputs is randomized, and the output of the model is the probability of the repeated reading trial being second in order.

\subsection{Baselines}
\label{sec:baselines}
\begin{itemize}[leftmargin=*]
    \item \textbf{Majority Class}: the most frequent class in the training set. As our data is balanced, this baseline is equivalent to a random choice.
    \item \textbf{Reading Speed}: the number of words read per second. Note that when the text is available, this measure can be calculated from the total reading time of the trial, and therefore does not require eye tracking. Prior work has consistently shown that reading is faster in repeated reading compared to first reading (see \Cref{sec:related-work}). We therefore expect this to be a strong baseline, which crucially enables determining the added value of eye tracking information for our decoding tasks.
\end{itemize}

\section{Experimental Setup}

\begin{table*}[ht]
\centering
\small
\resizebox{\textwidth}{!}{%
\begin{tabular}{llcccccc}
\toprule
\makecell{\textbf{Task}\\\textbf{Variant}} & 
\makecell{\textbf{Model}} & 
\makecell{\textbf{Eye Movements}\\\textbf{Input}} & 
\makecell{\textbf{New Item}\\\textbf{Seen Participant}} & 
\makecell{\textbf{New Participant}\\\textbf{Seen Item}} & 
\makecell{\textbf{New Item \&}\\\textbf{Participant}} & 
\multicolumn{2}{c}{\makecell{\textbf{All}}} \\
\midrule
\multirow{8}{*}{\makecell{Single\\Trial}} &
    Majority Class &
     - &
    $50.0 \color{white}|_{\pm0.0}$ &
    $50.0 \color{white}|_{\pm0.0}$ &
    $50.0 \color{white}|_{\pm0.0}$ &
    $50.0 \color{white}|_{\pm0.0}$ & 
    -  \\ \cmidrule{2-8} 

  & \multirow{2}{*}{Reading Speed} &
  $E_S^{W,r}$ &
  $66.9_{\pm2.0}$ &
  $67.1_{\pm2.1}$ &
  ${66.8_{\pm2.1}}$ &
  $66.6_{\pm1.2}$ & - \\
& 
  & 
  $E_{EZ}^{W,1} , E_S^{W,r}$ &
  ${67.3_{\pm2.2}}$ &
  ${67.3_{\pm2.0}}$ &
  $66.5_{\pm2.1}$ &
  ${67.1_{\pm1.2}}$ & n.s \\ \cmidrule{2-8} 
&
  \multirow{2}{*}{XGBoost} &
  $E_S^{W,r}$ &
  $69.5_{\pm2.0}$ &
  $70.7_{\pm2.0}$ &
  $68.7_{\pm2.0}$ &
  $69.6_{\pm1.2}$ & ** \\ 
&
  & 
  $E_{EZ}^{W,1} , E_S^{W,r}$ &
  ${70.1_{\pm2.1}}$ &
  ${\mathbf{71.2_{\pm1.9}}}$ &
  ${69.3_{\pm2.0}}$ &
  ${\mathbf{70.2_{\pm1.2}}}$ & *** \\ \cmidrule{2-8} 
&
  \multirow{2}{*}{RoBERTEye-Fixations} &
  $E_S^{W,r}$ &
  ${69.8_{\pm2.0}}$ &
  ${70.1_{\pm2.0}}$ &
  ${\mathbf{69.8_{\pm2.0}}}$ &
  ${69.9_{\pm1.1}}$ & * \\ 
&
  & 
  $E_{EZ}^{W,1} , E_S^{W,r}$ &
  $69.6_{\pm2.0}$ &
  $69.3_{\pm2.0}$ &
  $68.4_{\pm2.1}$ &
  $69.1_{\pm1.2}$ & * \\ \cmidrule{2-8} 
&
  \multirow{2}{*}{RoBERTEye-Words} &
  $E_S^{W,r}$ &
  $69.7_{\pm2.0}$ &
  ${71.0_{\pm2.0}}$ &
  $69.0_{\pm2.0}$ &
  ${69.9_{\pm1.1}}$ & ***\\ 
&
  & 
  $E_{EZ}^{W,1} , E_S^{W,r}$ &
  ${\mathbf{70.3_{\pm2.0}}}$ &
  $69.7_{\pm2.0}$ &
  ${69.7_{\pm2.0}}$ &
  ${69.9_{\pm1.2}}$ & ***\\ \midrule
  \multirow{4}{*}{\makecell{Paired\\Trials}} &
    Majority Class &
     - &
    $50.0 \color{white}|_{\pm0.0}$ &
    $50.0 \color{white}|_{\pm0.0}$ &
    $50.0 \color{white}|_{\pm0.0}$ &
    $50.0 \color{white}|_{\pm0.0}$ & 
    -  \\ \cmidrule{2-8} 
  & Reading Speed &
  \multirow{4}{*}{$E_S^{W,r}, E_S^{W,r'}$} &
  $88.0_{\pm2.0}$ &
  $88.1_{\pm2.1}$ &
  $87.2_{\pm2.1}$ &
  $87.7_{\pm1.2}$ & - \\
& 
  XGBoost &
  & 
  $\mathbf{91.5_{\pm1.8}}$ &
  $\mathbf{92.2_{\pm1.7}}$ &
  $\mathbf{90.6_{\pm1.8}}$ &
  $\mathbf{91.4_{\pm1.3}}$ & *** \\
& 
  RoBERTEye-Fixations &
  & 
  $85.3_{\pm2.2}$ &
  $86.0_{\pm2.1}$ &
  $84.5_{\pm2.4}$ &
  $85.3_{\pm1.3}$ & n.s \\
& 
  RoBERTEye-Words &
  & 
  $88.6_{\pm1.9}$ &
  $89.4_{\pm1.9}$ &
  $88.6_{\pm1.9}$ &
  $88.8_{\pm1.3}$ & n.s \\

\bottomrule
\end{tabular}
}
\caption{Test accuracy aggregated across 10 cross-validation splits, with 95\% confidence intervals. $E_{S}^{W,r}$ and $E_{EZ}^{W,1}$ are human and E-Z Reader-synthesized  eye movements respectively. Differences in performance across models are tested using a linear mixed effects model. In R notation: $is\_correct \sim model + (model \mid participant) + (model \mid paragraph)$. 
Significant gains over the reading speed baseline in the All regime are marked with '*' $p < 0.05$, '**' $p < 0.01$ and '***' $p < 0.001$. Within each task and evaluation regime, the best-performing model is in bold.}
\label{tab:main_res_table}
\end{table*}

\subsubsection*{Evaluation Regimes}

We use 10-fold cross validation with three evaluation regimes:
\begin{itemize}[leftmargin=*]
 \item \textbf{New Participant} eye movement data is available for the given paragraph, but no prior data was collected for the participant.
 \item \textbf{New Item} prior eve movement data is available for the participant, but not for the paragraph.
 \item \textbf{New Participant and Item} no prior data is available for the participant nor for the paragraph.
 \item \textbf{All} the union of the above three regimes.
\end{itemize}

\subsubsection*{Data Splits}
To allow complete matching of participants across the first and repeated reading of each article, out of the 10 articles read by each participant during first reading, we use only the 2 articles that were read twice. 
Further, we leverage the counterbalancing properties of OneStop to obtain data splits that fulfill the following properties: 1)  the three test regimes are balanced in number of participants 2)  the three validation regimes are balanced to the extent possible in number of participants 3) there is an equal number of consecutive and non-consecutive repeated readings in each portion of the split. 

We define a constrained combinatorial problem that has an algorithmic solution that satisfies these constraints. We provide further details on the solution in \Cref{app:CV}. All resulting splits satisfy that the training set has 264 participant-article pairs, the validation set has 48 pairs, and the test set has 54 pairs, where each test regime has exactly 18 pairs, all balanced with respect to consecutive and non-consecutive repeated reading. In \Cref{fig:CV} we present an example of one split.

\subsubsection*{Model Training and Selection}
We perform hyperparameter optimization and model selection separately for each split. We assume that at test time, the evaluation regime of the trial is \emph{unknown}. Model selection is therefore based on the entire validation set of the split. 
All neural network-based models were trained using the PyTorch Lighting \hbox{\citep{Falcon_PyTorch_Lightning_2019}} library on L40S-48GB GPUs. Further details on the training procedure, including the full hyperparameter search space for all models are provided in \Cref{app:hardware-software} and \Cref{app:hyperparams}.


\section{Results}
\label{sec:results}
Below, we summarize our main experimental findings for both the paired and single-trial variants of the task. \Cref{tab:main_res_table} presents the quantitative results.

\paragraph{Single Trial} 
In this task all models outperform the reading speed baseline, demonstrating the added value of eye movement information.
The highest All Accuracy of $70.2$ is achieved with the XGBoost model augmented with \mbox{E-Z} Reader Scanpaths. However, it does not outperform the other models statistically, and different models come first on different evaluations, again, with no statistically significant differences from the other models. Within each model, performance is relatively stable across the three evaluation regimes. 

\paragraph{The Effects of Synthetic References} 

With one exception, the best performing model in each evaluation regime includes a synthetic scanpath augmentation. However, the gains over the non-augmented model counterparts are not statistically significant and not consistent within each model. 

\paragraph{Paired Trials} In the paired setup, the model’s output is an ordering of two trials. To make the evaluation of these predictions comparable to the single-trial task, we “unaggregate” the model’s predictions so that predicting the correct order counts as two correct single-trial classifications (and vice versa for an incorrect prediction).
The reading speed baseline achieves a high All Accuracy of $87.7$. While the neural models exhibit baseline-level results, XGBoost substantially outperforms the baseline and the neural models in all the evaluation regimes, reaching an overall Accuracy of $91.4$. Overall, the feature-based method tends to yield stronger results than the neural methods.

In \Cref{app:results} we present complementary evaluation measures such as Precision, Recall and F1 for both the validation and test partitions.

    

\section{Fine-Grained Analysis of Model Performance}
\label{sec:error-analysis}

The controlled experimental design of OneStop enables going beyond aggregated performance measures and understanding model behavior as a function of trial characteristics. Prior work with OneStop observed that individual eye movement measures in first and repeated reading vary systematically across different item and participant characteristics \citep{meiri2024deja}. Here, we analyze this variability from the perspective of the classification performance of models that integrate multiple eye movement features. This allows for a fine-grained characterization of model behavior with respect to different data characteristics, and further leverages the models as an analytic tool for inspecting the data itself. 

\begin{figure}
    \centering
    \includegraphics[width=1\linewidth]{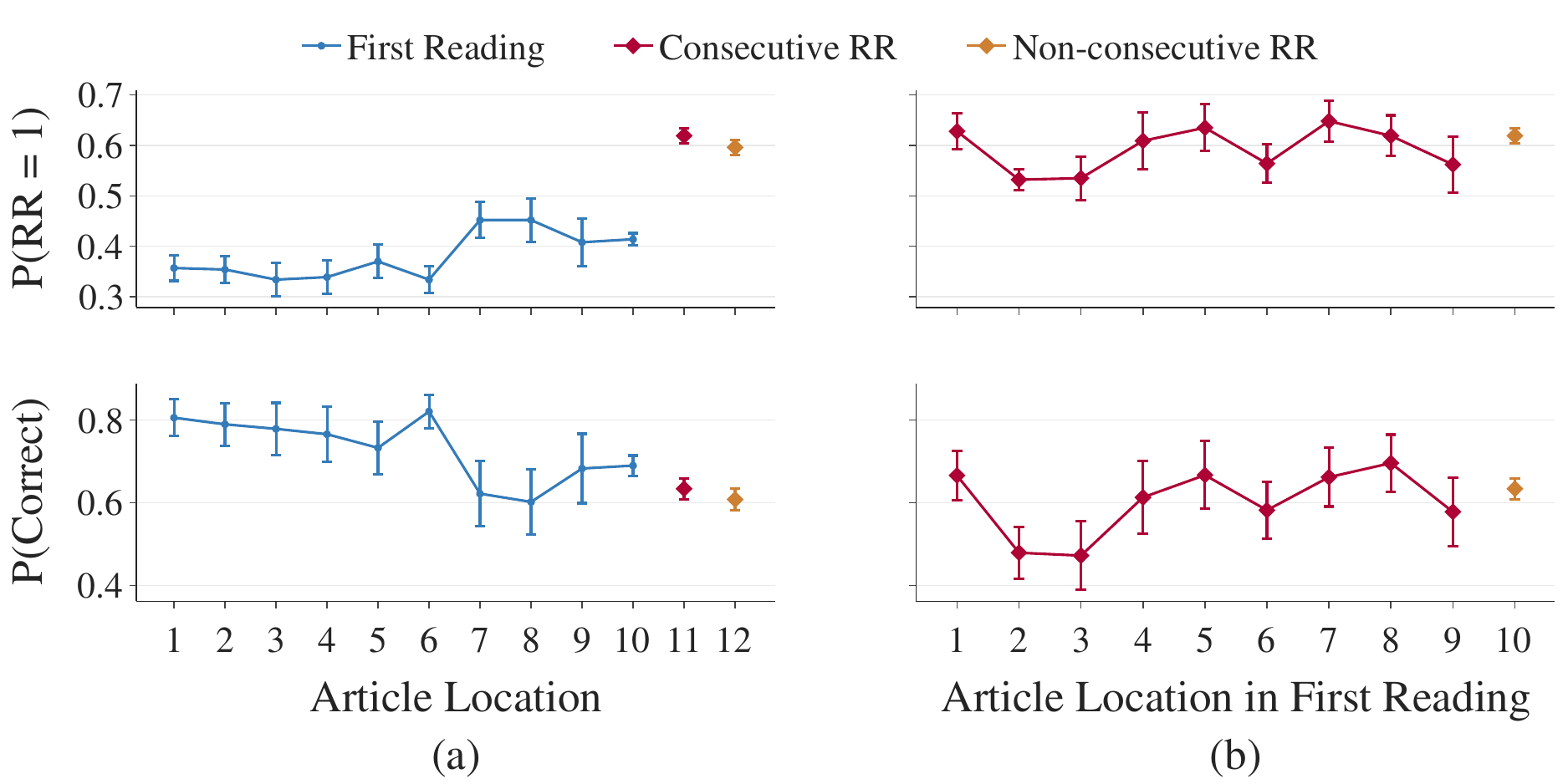}
    \caption{Analysis of the \mbox{E-Z} Reader augmented XGBoost model's behavior as a function of item position. Depicted are probability assignment (top) and classification accuracy (bottom) with 95\% confidence intervals. (a) First and repeated reading (RR) as a function of article position in the experiment. (b) Repeated reading as a function of the article position in the first reading. See \Cref{fig:exp-schema} for the experiment structure.}
    \label{fig:post-analysis-article-idx}
\end{figure}

Specifically, we focus on the more challenging single trial task, and analyze the assigned probabilities and prediction accuracy of the best performing model, XGBoost augmented with \mbox{E-Z} Reader scanpaths. 
\Cref{fig:post-analysis-article-idx} presents the mean probability assigned to trials for being a repeated reading trial (top) and the mean probability of classifying trials correctly (bottom) as a function of the position of the article in the experiment, see \Cref{fig:exp-schema}.


In \textbf{first reading} trials (blue) in \Cref{fig:post-analysis-article-idx} (a), as the experiment progresses, the model exhibits decreasing confidence in classifying trials as first reading ($p<10^{-8}$) and goes down in prediction accuracy ($p\approx0.001$).
This outcome mirrors a decrease in reading times during first reading as the experiment progresses observed in \citet{meiri2024deja}. When adding reading speed as a predictor in models that predict the probability assignments for repeated reading and accuracy\footnote{In R notation: $outcome \sim fixed\_terms + (fixed\_terms | subject) + (fixed\_terms | paragraph)$ where $outcome \in \{P(RR=1), P(Correct)\}$, and $fixed\_terms=position+reading\_speed$}, we find that for both the effect of position is no longer significant ($p\approx0.7$ for probabilities and $p\approx0.5$ for accuracy). This suggests that the model heavily relies on reading speed or correlates thereof. Consequently, as reading speed increases during the experimental session and comes closer to the reading speed in repeated reading, the model becomes worse at correctly classifying first reading items. 


In \textbf{repeated reading} trials in \Cref{fig:post-analysis-article-idx} (a), the model assigns higher probabilities to consecutive repeated readings compared to non-consecutive ones ($p\approx0.02$). Accuracy is also numerically lower compared to consecutive repeated readings, but the difference is not significant  ($p\approx0.16)$. These outcomes are again in line with the analysis of \citet{meiri2024deja} who found lower reading times and less skipping in non-consecutive versus consecutive reading. Taken together with these results, our analysis strengthens \citet{meiri2024deja} interpretation that reading facilitation is greater in consecutive reading potentially due to better memory retention of the first reading. 

In \Cref{fig:post-analysis-article-idx} (b) we examine repeated reading item probabilities and accuracy as a function of the position of the first reading. Reflecting again reading speed and other single measure analyses in \citet{meiri2024deja}, we find no evidence for a better model's classification accuracy with fewer articles between the two readings ($p\approx0.16$). However, we do find an effect in model probabilities for repeated reading which increases with article position ($p\approx0.03$), hinting at favorable effect of the recency of the first reading. This suggests that model analysis can unveil finer grained patterns in the data than with traditional univariate analyses.


\section{Conclusion and Discussion}
\label{sec:discussion}

Our study presents the first attempt to decode the number of prior interactions of a reader with a text from their eye movements. We demonstrate that it is feasible to perform this task, with various degrees of success depending on the difficulty of the task variant. In addition, we propose and experiment with a general method for leveraging synthetic ordinary reading data to improve predictive modeling of non-ordinary reading. Overall, the results indicate that there is highly informative signal for the presence or absence of a prior text interaction in eye movements at the level of a single paragraph and reader. This signal tends to be better captured by feature-based models than by neural language models. This outcome provides further evidence for the informativeness of eye movements regarding the reader's cognitive state during online language processing. It also opens the door for practical, user facing applications in areas such as education and online content delivery, where inferences about the current interaction of the reader with the text can be used for customizing and personalizing upcoming textual content to best support user goals.

\section{Ethical Considerations}

The eyetracking data used in our experiments was collected under an institutional IRB protocol \cite{onestop2025preprint}. All the participants provided written consent prior to taking part in the eyetracking experiment and received monetary compensation for their participation. The dataset is anonymized. Analyses and modeling of eye movements in repeated reading are among the main use cases for which the data was collected.

Decoding of repeated reading can be valuable in applications for monitoring reading acquisition, learning progress and effectiveness of retaining learned material. It can further facilitate special assistance to individuals and populations that struggle with reading comprehension, which can potentially be diagnosed via repeated reading. However, such technologies also pose a potential for inaccurate predictions and biases that may put various individuals and populations at a disadvantage. These include L2 learners, participants with cognitive impairments, participants with eye conditions and others. Additional data collection, modeling and analysis work for these groups is required before considering the deployment of such technology.

Finally, it is important to consider the issues of privacy and consent in the scope of eyetracking technologies. It was previously shown that eye movements contain information that can be used for user identification \cite[e.g.][]{bednarik2005eye,jager2020deep}. We do not perform user identification in this study, and point out the importance of not storing information that could enable participant identification in future studies on repeated reading and other reading regimes. We further stress that future systems that perform prediction of repeated reading are to be used only with explicit consent from potential users to have their eye movements collected and analyzed for this purpose.

\section{Limitations}
\label{sec:limitations}

Our work has a number of limitations that are related to the experimental design and the eyetracking data. Consecutive repeated reading occurs at the level of a full article, such that there are minimally 3 intervening paragraphs between two readings of the same paragraph. This setup does not address immediate repeated reading that involves working memory. The maximal amount of intervening material between two readings of the same article is 10 articles, leaving out larger time intervals between the readings. The experiment is also restricted to two readings of any given text, while in daily life the same text can be read more than twice. The underlying texts are all newswire articles, and while they include a wide range of topics, other textual domains are not covered. We intend to collect data and investigate both shorter and longer repetition intervals, multiple repeated readings and additional textual domains in future work.

An additional limitation is the experimental procedure, where reading occurs in-lab, and the presence of a reading comprehension question after each paragraph. Both aspects can negatively affect the ecological validity of the data and lead to reading behavior that is not fully representative of everyday life. Relatedly, while we use the term ordinary reading to refer to reading for general text comprehension, we acknowledge that this term, and similar terms such as ``normal reading'', are not without faults \cite{huettig2022myth}. In information seeking, while question answering is a general framework for formulating information seeking tasks, the experiment captures only a limited range of tasks that are restricted to a single paragraph and constrained by the annotation structure of STARC. Furthermore, readers who forgot the task cannot return to it during paragraph reading. Future work with different tasks, amounts of text, and experimental setups is needed to address these limitations.

Additional limitations concern the participants, the experiment language and the equipment used. Although OneStop \citep{onestop2025preprint} is the first public dataset that enables studying repeated reading, it is restricted to adult L1 speakers, with no cognitive impairments, and mostly with no eye conditions. This pool of participants does not cover multiple populations, including L2 speakers, children, elderly, participants with cognitive and physical impairments and others. 
Moreover, the eyetracking data and modeling work is restricted to English. 
These factors limit the scope and the generality of the results. Both data collection and model development work is required to include additional languages and populations. Finally, our approach has currently only been tested using a state-of-the-art eyetracker (Eyelink 1000 Plus) at a sampling rate of 1000Hz. This eye tracker allows extracting gaze position and duration at a very high temporal resolution and character-level precision. 
Such equipment is generally not available for end users, limiting the application potential of the current work. The feasibility of using lower spatial and temporal resolution eye tracking systems, as well as standard front-facing cameras on devices such as laptops, tablets and phones should be examined in future work.

\bibliography{custom}

\clearpage

\appendix

\section{Word and Fixation Level Features}
\label{app:word_fix_level_features}

This section describes the features that constitute both \( e^{word} \) and \( e^{fix} \) in human and generated trials. The full lists of eye movement features at both word and fixation levels appear in \Cref{app:tab:combined-eye-movement-features}. Additionally, the linguistic word properties that, together with word-level eye movement features, constitute \( e^{word} \) are listed in \Cref{app:tab:linguistic-features}.

Here we provide definitions of the six standard eye movement measures presented in \Cref{sec:feature_based_methods}:

\paragraph{Total Fixation Duration (TFD)} The total duration of fixations on a word.

\paragraph{Gaze Duration (GD)} The total duration of all fixations on a word from the first time entering it to the first time exiting it. See \Cref{app:synthesis_details} for details on GD implementation in \mbox{E-Z} Reader.

\paragraph{First Fixation Duration (FFD)} The duration of the first fixation on a word.

\paragraph{Regression Rate} The proportion of saccades per word that move regressively (backwards).

\paragraph{Skip Rate} Unlike the other features defined in this section, this is a trial-level feature, defined as the proportion of skipped words in a trial (i.e., the fraction of words where \texttt{total\_skip = 1}; see \Cref{app:tab:combined-eye-movement-features}).

\begin{table*}[ht]
\centering
\resizebox{\textwidth}{!}{%
\begin{tabular}{@{}ll@{}}
\toprule
\multicolumn{1}{c}{\multirow{2}{*}{\textbf{Feature Name}}} & \multicolumn{1}{c}{\multirow{2}{*}{\textbf{Description}}}                                                  \\
\multicolumn{1}{c}{}                                       & \multicolumn{1}{c}{}                                                                                       \\ \midrule
\textbf{Word-Level Eye Movement Features}                  &                                                                                                            \\ \midrule
IA\_DWELL\_TIME                                            & (\textbf{TFD}) The sum of the duration across all fixations that fell in the current interest area                        \\
IA\_DWELL\_TIME\_\%                                        & Percentage of trial time spent on the current interest area (IA\_DWELL\_TIME / PARAGRAPH\_RT).        \\
IA\_FIXATION\_COUNT                                        & Total number of fixations falling in the interest area.                                                    \\
IA\_REGRESSION\_IN\_COUNT                                  & (\textbf{Regression Rate}) Number of times interest area was entered from a higher IA\_ID (from the right in English).                \\
IA\_REGRESSION\_OUT\_FULL\_COUNT                           & Number of times interest area was exited to a lower IA\_ID (to the left in English).                       \\
IA\_FIRST\_FIX\_PROGRESSIVE                                & Checks whether the first fixation in the interest area is a first-pass fixation.                           \\
IA\_FIRST\_FIXATION\_DURATION                              & (\textbf{FFD}) Duration of the first fixation event that was within the current interest area                             \\
IA\_FIRST\_RUN\_DWELL\_TIME & (\textbf{GD}) Dwell time of the first run (i.e., the sum of the duration of all fixations in the first run of fixations within the current interest area). \\
IA\_TOP                                                    & Y coordinate of the top of the interest area.                                                              \\
IA\_LEFT                                                   & X coordinate of the left-most part of the interest area.                                                   \\
normalized\_Word\_ID                                       & Position in the paragraph of the word interest area, normalized from zero to one.                          \\
IA\_REGRESSION\_OUT\_COUNT  & Number of times interest area was exited to a lower IA\_ID (to the left in English) before a higher IA\_ID was fixated in the trial.         \\
PARAGRAPH\_RT                                              & Reading time of the entire paragraph.                                                                      \\
total\_skip                                                & Binary indicator whether the word was fixated on.                                                          \\ \midrule
\textbf{Fixation-level Eye Movement Features}              &                                                                                                            \\ \midrule
CURRENT\_FIX\_INDEX                                        & The position of the current fixation in the trial.                                                         \\
CURRENT\_FIX\_DURATION                                     & Duration of the current fixation.                                                                          \\
CURRENT\_FIX\_X                                            & X coordinate of the current fixation.                                                                      \\
CURRENT\_FIX\_Y                                            & Y coordinate of the current fixation.                                                                      \\
CURRENT\_FIX\_INTEREST\_AREA\_INDEX                        & The word index (IA\_ID) on which the current fixation occurred.                                            \\
NEXT\_FIX\_INTEREST\_AREA\_INDEX                           & The word index (IA\_ID) on which the next fixation occurred.                                               \\ \bottomrule
\end{tabular}%
}
\caption{Word-level and fixation-level eye movement features, defined and extracted by SR Data Viewer.}
\label{app:tab:combined-eye-movement-features}
\end{table*}

\begin{table*}[ht]
\centering
\resizebox{\textwidth}{!}{%
\begin{tabular}{@{}ll@{}}
\toprule
\multicolumn{1}{c}{\textbf{Feature Name}} & \multicolumn{1}{c}{\textbf{Description}} \\
\midrule
Surprisal &  \makecell[l]{\cite{hale2001probabilistic,levy2008expectation}, formulated as $-\log_2(p(word|context))$ for each \textit{word} given the preceding textual content of the \\ paragraph as \textit{context}, probabilities extracted from the GPT-2-small language model \cite{radford2019language,wolf-etal-2020-transformers}.} \\
Wordfreq\_Frequency & Frequency of the word based on the Wordfreq package \cite{robyn_speer_2022_7199437}, formulated as $-\log_2(p(word))$. \\
Length & Length of the word in characters. \\
start\_of\_line & Binary indicator of whether the word appeared at the beginning of a line. \\
end\_of\_line & Binary indicator of whether the word appeared at the end of a line. \\
Is\_Content\_Word & \makecell[l]{Binary indicator of whether the word is a content word. \\ A content word is defined as a word that has a part-of-speech tag of either PROPN, NOUN, VERB, ADV, or ADJ.} \\
n\_Lefts & The number of leftward immediate children of the word in the syntactic dependency parse.\\
n\_Rights & The number of rightward immediate children of the word in the syntactic dependency parse.
\\
Distance2Head & The number of words to the syntactic head of the word.\\
\bottomrule
\end{tabular}%
}
\caption{Linguistic word properties and their descriptions. POS tags and parse trees were obtained using SpaCy \cite{Honnibal2020}.}
\label{app:tab:linguistic-features}
\end{table*}

\section{Trial Level Feature Sets}
\label{app:trial_level_features}

For all tasks, we computed all features for each trial independently of other trials.

In all feature-based models, we address the strong co-linearity observed among several features by applying Principal Component Analysis (PCA). A PCA model is fit on the training set, and the training, validation, and test features are transformed using the trained PCA. The number of PCA components is determined as the minimum required to maintain a specified fraction of explained variance. This fraction is optimized during hyperparameter tuning (see \Cref{app:hyperparams} for the search space we use).

In addition to the six standard eye movement features listed in \Cref{sec:feature_based_methods}, we also computed two additional measures for each trial:
\begin{itemize}
    \item \textbf{\texttt{num\_of\_words\_with\_TFD\_GD\_diff}:} This is the proportion of fixated words for which TFD $>$ GD, indicating refixations on a word after the first pass.
    
    \item \textbf{\texttt{mean\_without\_first\_run\_dwell\_time}:} For words fixated more than once (including first pass only), this feature represents the average extra fixation duration per additional fixation (i.e., TFD minus GD, divided by the number of additional fixations). If no word is fixated more than once, the value is set to 0.
\end{itemize}


\subsection{Word Property Coefficients}
\label{app:response_to_ling}
The formula for the linear model is:
\begin{equation*}
\begin{aligned}
Measure \sim 1 + Surp + Freq + Length + \\
Freq:Length + normalized\_word\_index
\end{aligned}
\end{equation*}

For each trial, we fit a linear model using the $OLS$ function from the \texttt{Statsmodels} library \cite{seabold2010statsmodels}. Before fitting the model, we normalize all measures. In order to maintain the assumptions of the linear model, we exclude zero values for the measures TFD, FFD, GD (their original distribution is normally-shaped with a point mass at zero due to the high number of skips).
Surprisal \cite{hale2001probabilistic,levy2008expectation} is defined as $-\log_2(p(word|context))$ for each \textit{word} given the preceding textual content of the 
 textual item as \textit{context}, probabilities extracted from the GPT-2-small language model \cite{radford2019language,wolf-etal-2020-transformers}. Frequency is based on the Wordfreq package \cite{robyn_speer_2022_7199437}, formulated as $\log_2(p(word))$. Length is defined by the number of characters, ignoring punctuation.
 We also include $normalized\_word\_index$ following the results presented in \cite{shubi2023}, which show general decrease in reading times for later words within each paragraph in OneStop.

\subsection{Saccade Network Measures}
\label{app:network_measures}

As described in \Cref{sec:feature_based_methods}, we define the directed graph $G=\{V,T\}$ such that $V$ is the set of words in $W$, and for all $u,v\in V$:
 $$T=\{\left(u,v\right): \text{there is a saccade from }u \text{ to }v\}$$ 

\begin{figure*}
    \centering
    \includegraphics[width=1\linewidth]{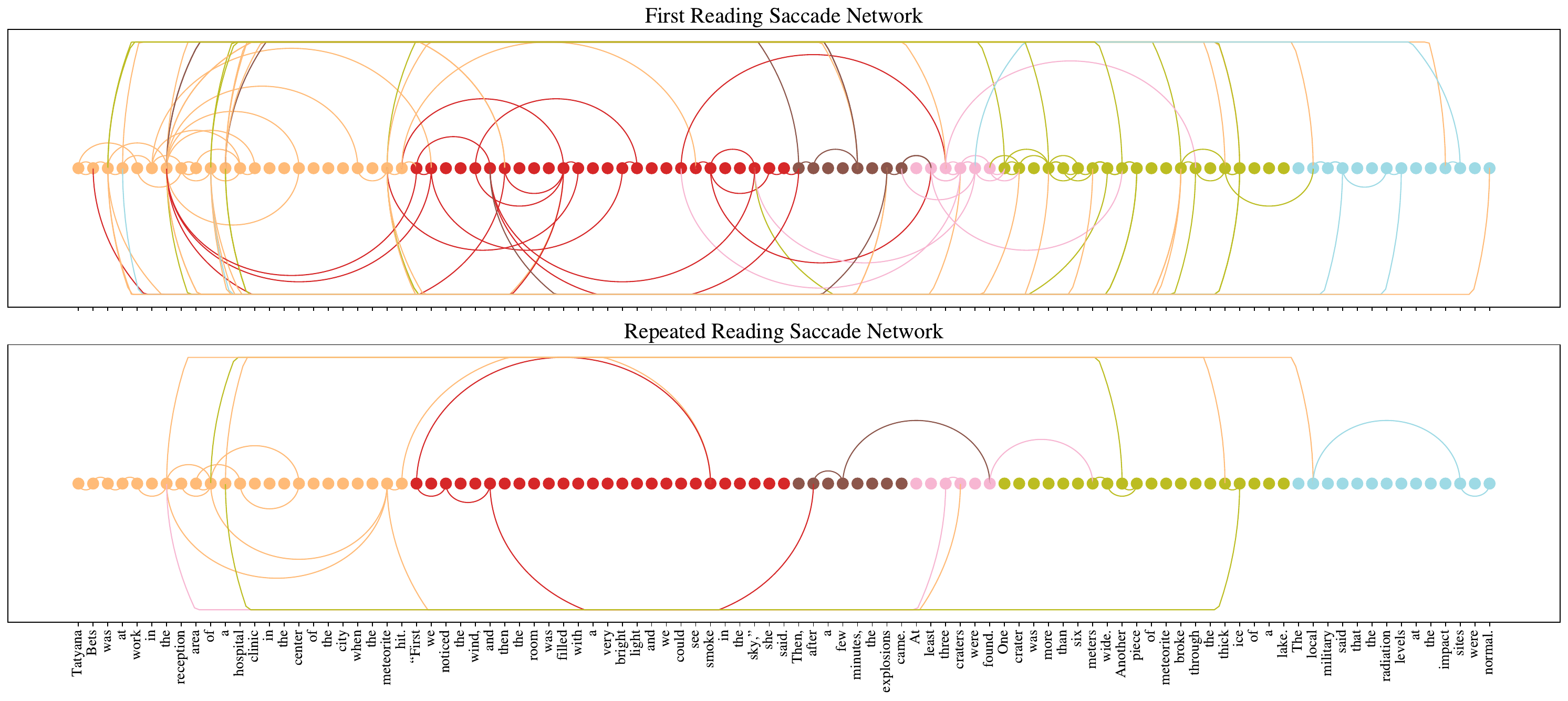}
    \caption{Visualization of two Saccade Networks as defined in \Cref{sec:feature_based_methods}.The top network represents the first reading, while the bottom network corresponds to the repeated reading of the same paragraph by the same participant. Different colors indicate different sentences within the paragraph.}
    \label{fig:sacc-net-viz}
\end{figure*}

A visualization example of two such networks appears in \Cref{fig:sacc-net-viz}.
 
 The following measures are computed for each saccade network instance:
\begin{enumerate}
    \item \textbf{Average Degree}:
    \[
    \text{Avg Degree} = \frac{\sum_{v \in V} \deg(v)}{|V|}
    \]
    where \( \deg(v) \) is the degree of vertex \( v \), and \( |V| \) is the number of vertices in \( G \).

    \item \textbf{Density}:
    \[
    \text{Density} = \frac{2|T|}{|V|(|V| - 1)}
    \]
    for an undirected graph \( G \), where \( |T| \) is the number of edges in \( G \).

    \item \textbf{Average Clustering Coefficient}:
    \[
    \text{Avg CC} = \frac{1}{|V|} \sum_{v \in V} C(v)
    \]
    where \( C(v) \) is the clustering coefficient of vertex \( v \), defined as the fraction of pairs of neighbors of \( v \) that are connected.

    \item \textbf{Average Betweenness Centrality}:
    \[
    \text{Avg Betweenness} = \frac{1}{|V|} \sum_{v \in V} b(v)
    \]
    where \( b(v) \) is the betweenness centrality of vertex \( v \), defined as the fraction of shortest paths in \( G \) that pass through \( v \).

    \item \textbf{Average Closeness Centrality}:
    \[
    \text{Avg Closeness} = \frac{1}{|V|} \sum_{v \in V} c(v)
    \]
    where \( c(v) \) is the closeness centrality of vertex \( v \), defined as the reciprocal of the average shortest path length from \( v \) to all other vertices in \( G \).

    \item \textbf{Transitivity}:
    \[
    \text{Transitivity} = \frac{3 \times \text{number of triangles}}{\text{number of connected triplets}}
    \]
    where a triangle is a set of three mutually connected vertices, and a triplet is a set of three vertices where at least two are connected.

    \item \textbf{Number of Bridges}:
    \begin{equation*}
\begin{aligned}
    \text{num\_bridges} = |\{e \in T \mid G \setminus \{e\} \\ \text{ has more connected components than } G\}|
\end{aligned}
\end{equation*}
    where a bridge is an edge whose removal increases the number of connected components in \( G \).
\end{enumerate}

\section{Scanpath Generation}
\label{app:synthesis_details}
For eye movement data generation we use \mbox{E-Z} Reader 10.4 \cite{veldre2023understanding} with the default parameters estimated from \cite{schilling1998comparing}:
\begin{align*}
    &A = 25.0, \quad \alpha_1 = 124.0, \quad \alpha_2 = 11.1,\\
    &\alpha_3 = 76.0, \quad \delta = 1.68, \quad \epsilon_1 = 0.1,\\
    &\epsilon_2 = 0.5, \quad \epsilon_3 = 1.0, \quad \eta_1 = 0.5, \quad \xi = 0.5\\
    &\eta_2 = 0.1, \quad I = 50.0, \quad \lambda = 0.25,\\
    &M_1 = 150.0, \quad M_2 = 25.0, \quad \omega_1 = 6.0,\\
    &\omega_2 = 3.0, \quad p_F = 0.01, \quad \psi = 7.0,\\
    &S = 25.0, \quad \sigma_{\gamma} = 20.0, \quad V = 60.0
\end{align*}
Parameter definitions appear in \cite{reichle_using_2013}. We set the  \texttt{includeRegressionTrials} parameter to \texttt{True} to allow inter-word regressions Out of the 1000 generated scanpaths per text, we choose the prototype scanpath to be the one which minimizes the mean Scasim \cite{von2011scanpath} distance to all other scanpaths. We use Scasim with the following formula 
\begin{align*}
    &\texttt{CURRENT\_FIX\_DURATION} \sim
    \texttt{CURRENT\_FIX\_X}\\
    & + \texttt{CURRENT\_FIX\_Y}
\end{align*}
with the parameters \texttt{center\_x=1280}, \texttt{center\_y=720}, \texttt{distance=77}, \texttt{unit\_size=1/60} and \texttt{normalize=False} because all scanpaths correspond to the same text and model parameters.

\subsection{Modifications in Word and Fixation Level Features}
For each text, we compute the same set of word- and fixation-level features used in human trials, as detailed in \Cref{app:word_fix_level_features}. Here, we highlight differences in feature definitions between human trials and those generated by the \mbox{E-Z} Reader model.

In OneStop, each trial corresponds to a single reading of a textual item by one subject. As a result, the word-level measures for a trial are extracted from a single scanpath. In contrast, the E-Z Reader model derives its word-level measures by aggregating scanpaths from 1000 statistical subjects. This aggregation introduces several differences compared to human-derived features:

\begin{itemize}
    \item \textbf{\texttt{IA\_SKIP}:} \\
    In human trials (see \Cref{app:tab:combined-eye-movement-features}), \texttt{IA\_SKIP} is binary, indicating whether a word was skipped. In E-Z Reader trials, however, this variable takes on a continuous value between 0 and 1, representing the proportion of statistical subjects who skipped the word.
    
    \item \textbf{Reading Time Measures for Skipped Words:} \\
    For human data, skipped words (i.e., those with \texttt{total\_skip = 1}) are assigned a value of zero for all reading-time features (e.g., TF, GD, FFD). In the E-Z Reader model, reading-time measures are summed over all statistical subjects and then normalized by the number of subjects who fixated the word. Consequently, these measures reflect only the fixation cases. To address this discrepancy, we excluded non-fixated words from human trials in the analysis presented in \Cref{tab:gen-quality} and from the trial-level feature extraction used in the feature-based models.
    
    \item \textbf{\texttt{PARAGRAPH\_RT} and \texttt{IA\_DWELL\_TIME\_\%}:} \\
    For E-Z Reader trials, these measures were computed using an "expected \texttt{IA\_DWELL\_TIME}," which is calculated as
    \[
    \texttt{IA\_DWELL\_TIME} \times \left(1 - \texttt{total\_skip}\right).
    \]
    This adjustment ensures that the measures account for the proportion of fixated versus skipped words.
\end{itemize}

\paragraph{Gaze Duration (GD)} We retain the original implementation of GD, which differs from the version provided in SR Data Viewer. In \mbox{E-Z} Reader, GD is computed at the word level as: $$GD(w)=\frac{\sum_{S_{FP}}{\text{First run dwell time}}}{|S|}$$
Where $S$ represents the subset of statistical subjects who fixated on word $w$, and $S_{FP}$ represents the subset of statistical subjects who fixated on word $w$ \textbf{during first pass}.
This formulation can result in GD being smaller than FFD in some cases.

\paragraph{Fixation Location on Screen} For human trials, the features \texttt{CURRENT\_FIX\_X} and \texttt{CURRENT\_FIX\_Y} specify the coordinates of each fixation on the screen. As \mbox{E-Z} Reader is inherently incapable of providing such features, we approximate them by using the center of each word

\subsection{Synthetic Data Analysis}
For the comparison between human and \mbox{E-Z} Reader generated trials, presented in \Cref{tab:gen-quality}, we use the same set of trial pairs as in model training (both first reading and repeated reading). For each human trial, we first extract all measures listed in \Cref{tab:gen-quality}, yielding a single value per measure type and human trial (in total: 4 measures $\times$ (1944 $\times$ 2) trials).
To obtain the values presented in \Cref{tab:gen-quality}, for each combination of eye movement measure and comparison type (either First Reading versus \mbox{E-Z} Reader or Repeated Reading versus \mbox{E-Z} Reader) we fit a linear mixed model formulated as $measure\_diff \sim 1 + (1|text)$, and extract the mean and standard error of the intercept.

\begin{figure*}
    \centering
    \includegraphics[width=\linewidth]{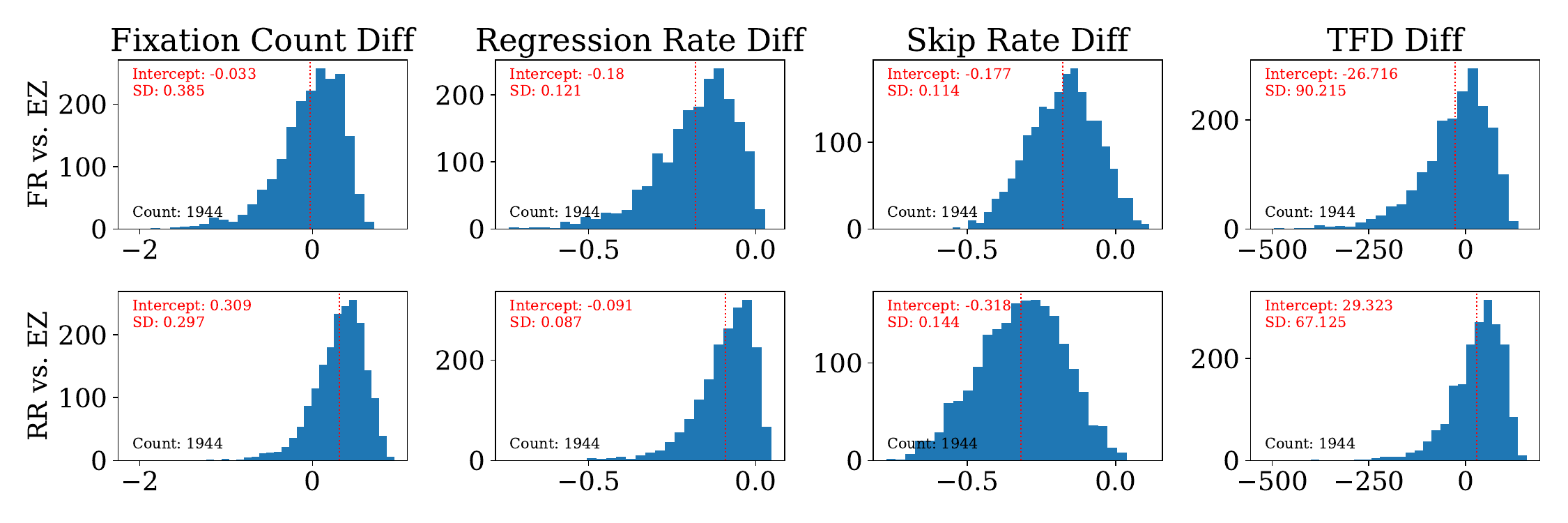}
    \caption{This visualization displays histograms representing the trial-level differences between human trials and the corresponding E-Z Reader synthesized trials. The arrangement of histograms mirrors the transposed rows and columns of \Cref{tab:gen-quality}. In each histogram, the top left corner shows the mean and standard deviation of the values.}
    \label{app:fig:gen-quality}
\end{figure*}

For comparing the absolute differences between first, repeated reading and \mbox{E-Z} Reader, we apply the following procedure:

\begin{enumerate}[leftmargin=*, label=\textbf{\arabic*.}]
    \item \textbf{Modeling:} For each measure (Fixation Count, Mean TFD, Regression Rate, and Skip Rate), we fit a mixed-effects model:
    \[
    \resizebox{0.45\textwidth}{!}{$
    \texttt{measure\_diff} \sim \texttt{comparison\_type} + \left(1\mid \texttt{text\_id}\right)
    $}
    \]
    where \texttt{comparison\_type} is a binary indicator being 1 for first reading minus \mbox{E-Z} Reader differences and 0 for repeated reading minus \mbox{E-Z} Reader differences.

    \item \textbf{Test Statistic:} We define
    \[
    d=|\beta_{FR-EZ}|-|\beta_{FR-EZ}+\beta_{RR-EZ}|,
    \]
    where $\beta_{FR-EZ}$ is the intercept (first reading) and $\beta_{FR-EZ}+\beta_{RR-EZ}$ is the mean for \mbox{E-Z} Reader vs. repeated reading differences.
    
    \item \textbf{Bootstrapping:} We perform 1000 text-level bootstrap iterations (sampling with replacement) to derive the sampling distribution of $d$ and compute one-sided $p$-values for all measures other than Mean TFD (for which we compute two-sided).
\end{enumerate}
In addition, in \Cref{app:fig:gen-quality}, we provide histograms that illustrate the distribution of trial-level differences for each measure.

\label{app:sec:synth-data-analysis}

\section{Cross-Validation and Participant-to-Article Assignment}
\label{app:CV}

For our cross-validation procedure, we split the data separately for each combination of article batch (1–3) and reading regime (information-seeking, ordinary reading). Each such combination consists of 10 articles and 60 participants, resulting in 60 “article-participant” pairs including both first and repeated reading version of each aritlce-participant pair. 

\paragraph{Data Splitting Rationale.}
Our data splitting is derived from an assignment of 6 participants to each article (for a total of 60 participants). This assignment must satisfy the following:
\begin{enumerate}
    \item All 6 participants assigned to article~\(i\) have reread it (repeated reading).
    \item Of these 6 participants, exactly half (3) performed \emph{consecutive} rereading and the other half performed \emph{nonconsecutive} rereading.
    \item All 6 participants assigned to article~\(i\) have read different \emph{other} articles in the repeated reading. 
\end{enumerate}
Section~\ref{app:CV_alg_section} below details the algorithmic procedure used to achieve this participant-to-article assignment.

\paragraph{Illustration of the Splits.}
Figure~\ref{fig:CV} illustrates one of the cross-validation folds for a single combination of article batch and reading regime (60 participants out of the 180). Participants are grouped by columns (from \(6\cdot(i-1)\) to \(6 \cdot i\)) according to the article to which they are assigned. Each table entry corresponds to the first and repeated readings of article~\(i\) by participant~\(j\). Since each participant reads exactly \emph{two} articles in repeated reading, each column has exactly two non-empty entries (one for consecutive and one for nonconsecutive repeated reading).

To create the data splits for the \(i\)-th fold, we select article \((10-i)\) and its corresponding participants as the unseen item and participants for the validation split, and \((10+1-i)\) as the unseen item and participants for the test split. Figure~\ref{fig:CV} demonstrates this process for fold~1.

\paragraph{Balanced Evaluation Regimes.}
Thanks to the properties of our participant-to-article assignment, we ensure:
\begin{enumerate}
    \item \emph{Balanced evaluation regimes for validation and test sets.} For each set, we balance “new item,” “new participant,” and “new item \& participant” splits. Specifically, in the test set, each regime has 6 “article readings.” In the validation set, the new item and new participant splits have 5 “article readings” each, while the new item \& participant split has 6.
    \item \emph{Balanced rereading conditions.} In the test split, all evaluation regimes are further balanced with respect to consecutive and nonconsecutive repeated reading (3 of each per regime).
\end{enumerate}

\subsection{Participant-to-Article Assignment Algorithm}
\label{app:CV_alg_section}

To assign participants to articles under the constraints described above, we used \texttt{cvxpy} \cite{diamond2016cvxpy} to solve the optimization problem detailed in Algorithm~\ref{app:CV_alg}. This procedure is applied independently for each combination of article batch (3 batches) and reading regime (information-seeking, ordinary reading). For each combination, we extract matrices \(P\) and \(Q\) as defined in Algorithm~\ref{app:CV_alg}, then solve to obtain a feasible assignment of participants to articles.

The optimization problem includes several key constraints to ensure the assignment is valid and meets the desired balance:
\begin{itemize}
    \item \textbf{Constraint 1:} Each participant is assigned to exactly one article.
    \item \textbf{Constraints 2 and 3:} Guarantee that:
    \begin{enumerate}
        \item Exactly 6 participants are assigned to each article.
        \item All 6 assigned participants had reread that article.
        \item Among them, half had read it in a consecutive rereading and half in a nonconsecutive rereading.
    \end{enumerate}
    These constraints explain the “diagonal” structure in Figure~\ref{fig:CV} and the distribution of symbols '{\scriptsize \protect\divorced}' and '{\scriptsize \protect\unmarried}' within each diagonal cell.
    \item \textbf{Constraint 4:} Among participants assigned to article~\(i\), at most one participant has reread any other article. This maximizes the coverage of articles in the unseen participant regime and ensures exactly 5 different participants in the unseen participant regime for the validation split.
\end{itemize}

\begin{algorithm*}[ht!]
\caption{Participant-to-articles assignment algorithm}
\label{app:CV_alg}
\begin{algorithmic}
    \State \textbf{Input:} Parameters and variables:
    \begin{itemize}
        \item \(m\): number of items (articles).
        \item \(n\): number of participants.
        \item \(P \in \{0, 1\}^{m\times n}\): \(P_{i,j} = 1\) if participant \(j\) read article \(i\) in a \emph{consecutive} repeated reading, 0 otherwise.
        \item \(Q \in \{0, 1\}^{m\times n}\): \(Q_{i,j} = 1\) if participant \(j\) read article \(i\) in a \emph{nonconsecutive} repeated reading, 0 otherwise.
    \end{itemize}
    \State \textbf{Variable:} \( B \in \{0, 1\}^{n\times m} \)

    \State \textbf{Objective:} Minimize a constant (null optimization problem):
    \[
        \text{Minimize: } 0
    \]
    
    \State \textbf{Constraints:} 
    \begin{align*}
        & \text{Constraint 1: } \forall j \in [n]: \sum_{i\in [m]}{B_{j,i}} = 1 \\
        & \text{Constraint 2: } \forall i \in [m]: \text{row}_i(P) \cdot \text{col}_i(B) = 3 \\
        & \text{Constraint 3: } \forall i \in [m]: \text{row}_i(Q) \cdot \text{col}_i(B) = 3 \\
        & \text{Constraint 4: } (P+Q) \cdot B - 6\cdot I \leq 1
    \end{align*}
    
    \State \textbf{Output: } 
    Solve this constrained null optimization problem using \texttt{cvxpy} to obtain a feasible solution \(B\) which satisfies all constraints. The assignment for each participant \(j \in [n]\) is then
    \[
        \text{assigned\_article}(j) = \arg\max(\text{row}_j(B)).
    \]
\end{algorithmic}
\end{algorithm*}

This assignment directly underlies the cross-validation splits described earlier in this appendix (see Figure~\ref{fig:CV}).

\section{Hyperparameter Tuning}
\label{app:hyperparams}

We apply standardization for each feature in both word and fixation level representations. Mean and standard deviation are computed on the train set and applied to the validation and test sets, separately for each split. Feature normalization is performed using Scikit-learn \cite{pedregosa_scikit-learn_2011}.

For all the neural models, we use the AdamW optimizer \cite{loshchilov_decoupled_2018} with a batch size of $16$, a linear warmup ratio of $0.1$, and a weight decay of 0.1, following best practice recommendations from \citet{liu_roberta_2019} and \citet{mosbach_stability_2021}. The search space for learning rates is $\{0.00001, 0.00003, 0.0001\}$ and for dropout $\{0.1, 0.3, 0.5\}$. For RoBERTEye models, we allow the backbone RoBERTa weights to either be frozen or trainable. 
For all XGBoost based models, we searched over learning rate $\in \{0.3, 0.01, 0.001\}$, number of estimators $\in \{10, 100, 1000\}$, maximal tree depth $\in \{4, 6, 10\}$ and a regularization parameter $\alpha\in \{0, 0.1, 1, 10\}$. The XGBoost search space is a subset of the default search space used in \cite{h2o_Python_module} and includes the default hyperparameters implemented in \cite{chen2016xgboost}. In addition, in all feature based models we optimize the lower bound of the fraction of explained variance after a PCA transformation which constrains the number of components taken $\in \{0.8, 0.9, 1\}$.

\section{Hardware and Software}
\label{app:hardware-software}

All neural networks are trained using the Pytorch Lighting library \cite{Falcon_PyTorch_Lightning_2019,paszke2019pytorch} and evaluated using torch-metrics \cite{TorchMetrics_-_Measuring_2022} on a NVIDIA A100-40GB and A40-48GB GPUs.  We adapt Huggingface's RoBERTa implementation \cite{wolf-etal-2020-transformers}.
The baselines described in \Cref{sec:baselines} are reimplemented in this framework as well. A single training epoch took approximately 3 minutes. We train for a maximum of 30 epochs, stopping after 5 epochs without improvement on the validation set.

The number of model parameters for RoBERTEye is either between 2-3M parameters (depending on RoBERTEye-Words or RoBERTEye Fixations) when the backbone RoBERTa is frozen, otherwise 355M.

The code base for this project was developed with the assistance of GitHub Copilot, an AI-powered coding assistant. All generated code was carefully reviewed.

We utilized the \texttt{lme4} package in \texttt{R} \cite{Bates2015-ts} and the \texttt{MixedModels} package in \texttt{Julia} \cite{alday_2025_14838413} for fitting linear mixed-models.

\section{Data}
\label{sec:app-data}

We use OneStop Eye Movements \cite{onestop2025preprint} an eyetracking dataset in which native speakers of English read articles in English from the Guardian. The eyetracking data was collected with an Eyelink 1000 Plus eyetracker (SR Research) at a sampling rate of 1000Hz. The dataset includes 360 participants reading articles from the OneStopQA reading comprehension dataset \citep{berzak_starc_2020}, which includes 30 articles with 4-7 paragraphs (162 paragraphs in total). Each article is available in two text difficulty levels, an original Advanced version and a simplified Elementary version and each paragraph is accompanied by 3 multiple choice reading comprehension questions that can be answered based on any of the text difficulty versions. 

The 30 articles are divided into 3 batches. Each participant reads a single 10-article batch with a random ordering of the articles. Participants read each paragraph in one of its two difficulty versions on a separate screen, and then answer one of the three multiple choice reading comprehension questions on a new screen, without the ability to return to the paragraph. The participants are split between two reading regimes. Half of the participants are assigned to an ordinary reading regime, where the question is presented only after the paragraph. The other half is in an information-seeking regime where participants are also presented with the question (but not the answers) prior to reading the paragraph. In this work, we exclude all participants in the information seeking condition from both the analysis and the training procedure (180 participants are left).

After reading a 10-article batch, participants read two articles for a second time. In repeated reading, the paragraphs are identical to the first reading, while the questions are different. The article in position 11 is a \emph{consecutive} second presentation of the article in position 10. The article in position 12 is a \emph{nonconsecutive} second presentation of an article in one of the positions 1-9. Thus, half of the repeated reading data captures immediate consecutive rereading of the same article, and the other half is rereading with intervening reading material, ranging from 2 to 10 articles. Overall, there are 720 second presentations of articles, 360 in consecutive rereading in position 11 and 360 in a nonconsecutive rereading in position 12. The first reading of position 12 articles occurs 72 times in position 1 and 36 times in each of the positions 2-9. The 720 repeated article readings correspond to 3,888 paragraph trials with a total of 211,081 word tokens over which eye movements were collected, split equally between positions 11 and 12 and the two reading regimes.

\section{Results}
\label{app:results}
In this section, we present additional results for the two task variations. Validation accuracy is reported in \Cref{app:tab:val_accuracy}, F1 scores for both validation and test partitions are shown in \Cref{app:tab:f1_scores}, and Recall and Precision for both validation and test partitions are provided in \Cref{app:tab:prec_recall}. For all result tables, 95\% confidence intervals are standard normal bootstrap confidence interval \cite{davison1997bootstrap} with $B=1000$. In addition, when comparing between models, we also experimented with adding a random effect for the fold number, but the low variance between folds prevented the model from converging.

\renewcommand{\arraystretch}{1.5}
\begin{table*}[ht]
\centering
\small
\resizebox{\textwidth}{!}{%
\begin{tabular}{lllccccc}
\toprule
\makecell{\textbf{Eval}\\\textbf{Type}} & 
\makecell{\textbf{Task}\\\textbf{Variant}} &
\makecell{\textbf{Model}} &
\makecell{\textbf{Eye Movements}\\\textbf{Input}} &
\makecell{\textbf{New Item}\\\textbf{Seen Participant}} &
\makecell{\textbf{New Participant}\\\textbf{Seen Item}} &
\makecell{\textbf{New Item \&}\\\textbf{Participant}} &
\makecell{\textbf{All}} \\
\midrule
\multirow{12}{*}{\makecell{Validation}} & 
\multirow{4}{*}{\makecell{Paired\\Trials}} & 
Reading Speed & $E_S^{W,r}, E_S^{W,r'}$ & $87.7_{\pm2.3}$ & $88.8_{\pm2.2}$ & $87.1_{\pm2.1}$ & $87.9_{\pm1.3}$ \\[1ex]
 & & XGBoost & $E_S^{W,r}, E_S^{W,r'}$ & $92.4_{\pm1.8}$ & $92.5_{\pm1.8}$ & $90.7_{\pm1.9}$ & $91.8_{\pm1.0}$ \\[1ex]
 & & RoBERTEye Fixations & $E_S^{W,r}, E_S^{W,r'}$ & $87.4_{\pm2.2}$ & $89.8_{\pm2.0}$ & $87.9_{\pm2.0}$ & $88.3_{\pm1.2}$ \\[1ex]
 & & RoBERTEye Words & $E_S^{W,r}, E_S^{W,r'}$ & $92.4_{\pm1.8}$ & $92.4_{\pm1.7}$ & $90.8_{\pm1.8}$ & $91.8_{\pm1.0}$ \\ 
\cmidrule{2-8}
 & \multirow{8}{*}{\makecell{Single\\Trial}} & 
 \multirow{2}{*}{Reading Speed} & $E_S^{W,r}$ & $67.7_{\pm2.2}$ & $66.9_{\pm2.2}$ & $66.0_{\pm2.1}$ & $66.8_{\pm1.3}$ \\[1ex]
 & &  & $E_{EZ}^{W,1}, E_S^{W,r}$ & $66.7_{\pm2.2}$ & $66.8_{\pm2.2}$ & $66.5_{\pm2.2}$ & $66.7_{\pm1.3}$ \\ 
\cmidrule{3-8} 
 & & \multirow{2}{*}{XGBoost} & $E_S^{W,r}$ & $69.5_{\pm2.0}$ & $70.7_{\pm2.0}$ & $68.7_{\pm2.0}$ & $69.6_{\pm1.2}$ \\[1ex]
 & &  & $E_{EZ}^{W,1}, E_S^{W,r}$ & $70.1_{\pm2.1}$ & $71.2_{\pm1.9}$ & $69.3_{\pm2.0}$ & $70.2_{\pm1.2}$ \\ 
\cmidrule{3-8} 
 & & \multirow{2}{*}{RoBERTEye Fixations} & $E_S^{W,r}$ & $72.7_{\pm2.2}$ & $72.2_{\pm2.1}$ & $71.2_{\pm2.0}$ & $72.0_{\pm1.3}$ \\[1ex]
 & &  & $E_{EZ}^{W,1}, E_S^{W,r}$ & $72.8_{\pm2.1}$ & $73.0_{\pm2.1}$ & $71.3_{\pm2.0}$ & $72.3_{\pm1.2}$ \\ 
\cmidrule{3-8} 
 & & \multirow{2}{*}{RoBERTEye Words} & $E_S^{W,r}$ & $72.9_{\pm2.1}$ & $73.8_{\pm2.0}$ & $72.2_{\pm1.9}$ & $72.9_{\pm1.2}$ \\[1ex]
 & &  & $E_{EZ}^{W,1}, E_S^{W,r}$ & $72.5_{\pm2.2}$ & $72.7_{\pm2.1}$ & $70.8_{\pm2.0}$ & $72.0_{\pm1.2}$ \\ 
\bottomrule
\end{tabular}
}
\caption{Validation accuracy results for the two variants of the first vs. second reading prediction task with 95\% confidence intervals, aggregated across 10 cross-validation splits, and presented for both test and validation partitions.$E_{EZ}^{W,1}$ denotes synthesized eye movements generated using \mbox{E-Z} Reader \cite{reichle2003ez}.}
\label{app:tab:val_accuracy}
\end{table*}

\renewcommand{\arraystretch}{1.5}
\begin{table*}[ht]
\centering
\small
\resizebox{\textwidth}{!}{%
\begin{tabular}{lllc cccc}
\toprule
\multicolumn{1}{l}{\begin{tabular}[l]{@{}l@{}}Eval\\Type\end{tabular}} & 
\multicolumn{1}{l}{\begin{tabular}[l]{@{}l@{}}Task\\Variant\end{tabular}} &
\multicolumn{1}{l}{Model} &
\multicolumn{1}{c}{Eye Movements Input} &
\multicolumn{1}{c}{\begin{tabular}[c]{@{}c@{}}New Item\\ Seen Participant\end{tabular}} &
\multicolumn{1}{c}{\begin{tabular}[c]{@{}c@{}}New Participant\\ Seen Item\end{tabular}} &
\multicolumn{1}{c}{\begin{tabular}[c]{@{}c@{}}New Item \&\\ Participant\end{tabular}} &
\multicolumn{1}{c}{All} \\ \midrule
\multirow{12}{*}{Validation} & \multirow{4}{*}{Paired Trials} & 
Reading Speed & $E_S^{W,r}, E_S^{W,r'}$ & $87.9_{\pm2.4}$ & $88.5_{\pm2.3}$ & $87.3_{\pm2.2}$ & $87.9_{\pm1.3}$ \\ 
 &  & XGBoost & $E_S^{W,r}, E_S^{W,r'}$ & $92.5_{\pm1.9}$ & $92.4_{\pm1.8}$ & $90.7_{\pm2.0}$ & $91.8_{\pm1.1}$ \\ 
 &  & RoBERTEye Fixations & $E_S^{W,r}, E_S^{W,r'}$ & $87.9_{\pm2.3}$ & $89.8_{\pm2.1}$ & $88.1_{\pm2.2}$ & $88.6_{\pm1.3}$ \\ 
 &  & RoBERTEye Words & $E_S^{W,r}, E_S^{W,r'}$ & $92.5_{\pm1.8}$ & $92.4_{\pm1.8}$ & $90.9_{\pm1.8}$ & $91.9_{\pm1.1}$ \\ \cmidrule{2-8}
 & \multirow{8}{*}{Single Trial} & \multirow{2}{*}{Reading Speed} & $E_S^{W,r}$ & $65.1_{\pm2.7}$ & $64.7_{\pm2.7}$ & $65.3_{\pm2.5}$ & $65.1_{\pm1.5}$ \\ 
 &  &  & $E_{EZ}^{W,1}, E_S^{W,r}$ & $65.9_{\pm2.7}$ & $64.7_{\pm2.6}$ & $64.6_{\pm2.5}$ & $65.0_{\pm1.5}$ \\ \cmidrule{3-8} 
 &  & \multirow{2}{*}{XGBoost} & $E_S^{W,r}$ & $70.5_{\pm2.5}$ & $70.2_{\pm2.5}$ & $69.0_{\pm2.3}$ & $69.9_{\pm1.4}$ \\ 
 &  &  & $E_{EZ}^{W,1}, E_S^{W,r}$ & $70.4_{\pm2.5}$ & $68.3_{\pm2.6}$ & $68.8_{\pm2.5}$ & $69.1_{\pm1.5}$ \\ \cmidrule{3-8} 
 &  & \multirow{2}{*}{RoBERTEye Fixations} & $E_S^{W,r}$ & $71.3_{\pm2.6}$ & $71.1_{\pm2.6}$ & $70.3_{\pm2.4}$ & $70.9_{\pm1.4}$ \\ 
 &  &  & $E_{EZ}^{W,1}, E_S^{W,r}$ & $72.5_{\pm2.4}$ & $71.2_{\pm2.4}$ & $71.0_{\pm2.3}$ & $71.5_{\pm1.4}$ \\ \cmidrule{3-8} 
 &  & \multirow{2}{*}{RoBERTEye Words} & $E_S^{W,r}$ & $71.2_{\pm2.6}$ & $70.8_{\pm2.5}$ & $69.5_{\pm2.4}$ & $71.1_{\pm1.5}$ \\ 
 &  &  & $E_{EZ}^{W,1}, E_S^{W,r}$ & $71.1_{\pm2.5}$ & $71.7_{\pm2.5}$ & $70.5_{\pm2.3}$ & $71.1_{\pm1.5}$ \\ \midrule

\multirow{12}{*}{Test} & \multirow{4}{*}{Paired Trials} & 
Reading Speed & $E_S^{W,r}, E_S^{W,r'}$ & $87.8_{\pm2.2}$ & $87.9_{\pm2.2}$ & $87.2_{\pm2.1}$ & $87.6_{\pm1.3}$ \\ 
 &  & XGBoost & $E_S^{W,r}, E_S^{W,r'}$ & $91.5_{\pm1.8}$ & $92.1_{\pm1.8}$ & $90.6_{\pm1.9}$ & $91.4_{\pm1.1}$ \\ 
 &  & RoBERTEye Fixations & $E_S^{W,r}, E_S^{W,r'}$ & $85.6_{\pm2.4}$ & $86.2_{\pm2.3}$ & $85.5_{\pm2.3}$ & $85.7_{\pm1.3}$ \\ 
 &  & RoBERTEye Words & $E_S^{W,r}, E_S^{W,r'}$ & $88.6_{\pm2.1}$ & $89.4_{\pm2.0}$ & $88.7_{\pm2.1}$ & $88.9_{\pm1.2}$ \\ \cmidrule{2-8}
 & \multirow{8}{*}{Single Trial} & \multirow{2}{*}{Reading Speed} & $E_S^{W,r}$ & $64.8_{\pm2.4}$ & $65.3_{\pm2.6}$ & $65.2_{\pm2.5}$ & $65.1_{\pm1.4}$ \\ 
 &  &  & $E_{EZ}^{W,1}, E_S^{W,r}$ & $65.4_{\pm2.6}$ & $65.6_{\pm2.5}$ & $65.2_{\pm2.6}$ & $65.4_{\pm1.5}$ \\ \cmidrule{3-8} 
 &  & \multirow{2}{*}{XGBoost} & $E_S^{W,r}$ & $68.6_{\pm2.4}$ & $69.8_{\pm2.4}$ & $67.8_{\pm2.5}$ & $68.7_{\pm1.4}$ \\ 
 &  &  & $E_{EZ}^{W,1}, E_S^{W,r}$ & $68.7_{\pm2.4}$ & $70.3_{\pm2.3}$ & $68.6_{\pm2.4}$ & $69.2_{\pm1.4}$ \\ \cmidrule{3-8} 
 &  & \multirow{2}{*}{RoBERTEye Fixations} & $E_S^{W,r}$ & $68.2_{\pm2.4}$ & $68.2_{\pm2.5}$ & $68.5_{\pm2.4}$ & $68.3_{\pm1.4}$ \\ 
 &  &  & $E_{EZ}^{W,1}, E_S^{W,r}$ & $69.0_{\pm2.3}$ & $68.2_{\pm2.4}$ & $68.3_{\pm2.4}$ & $68.5_{\pm1.4}$ \\ \cmidrule{3-8} 
 &  & \multirow{2}{*}{RoBERTEye Words} & $E_S^{W,r}$ & $68.1_{\pm2.4}$ & $69.1_{\pm2.4}$ & $67.3_{\pm2.5}$ & $68.1_{\pm1.4}$ \\ 
 &  &  & $E_{EZ}^{W,1}, E_S^{W,r}$ & $68.0_{\pm2.5}$ & $67.4_{\pm2.5}$ & $67.7_{\pm2.5}$ & $67.7_{\pm1.5}$ \\ 
\bottomrule
\end{tabular}%
}
\caption{F1 results for the two variants of the first vs. second reading prediction task with 95\% confidence intervals, aggregated across 10 cross-validation splits, and presented for both test and validation partitions. $E_{EZ}^{W,1}$ denotes synthesized eye movements generated using \mbox{E-Z} Reader \cite{reichle2003ez}.}
\label{app:tab:f1_scores}
\end{table*}

\renewcommand{\arraystretch}{1.5}
\begin{table*}[ht]
\centering
\small
\resizebox{\textwidth}{!}{%
  \begin{tabular}{lllc*{4}{cc}}
    \toprule
    Eval Type & Task Variant & Model & Eye Movements Input & \multicolumn{2}{c}{New Item Seen Participant} & \multicolumn{2}{c}{New Participant Seen Item} & \multicolumn{2}{c}{New Item \& Participant} & \multicolumn{2}{c}{All} \\
    \cmidrule(lr){5-6}\cmidrule(lr){7-8}\cmidrule(lr){9-10}\cmidrule(lr){11-12}
              &              &       &                      & Prec. & Recall & Prec. & Recall & Prec. & Recall & Prec. & Recall \\
    \midrule
    \multirow{12}{*}{Validation} 
      & \multirow{4}{*}{Paired Trials} 
          & Reading Speed 
              & $E_S^{W,r},E_S^{W,r'}$ 
              & $87.1_{\pm3.2}$ & $88.7_{\pm3.1}$ 
              & $89.0_{\pm3.1}$ & $88.1_{\pm3.2}$ 
              & $87.5_{\pm2.9}$ & $87.2_{\pm2.9}$ 
              & $87.8_{\pm1.8}$ & $88.0_{\pm1.7}$ \\[1ex]
      & 
          & XGBoost 
              & $E_S^{W,r},E_S^{W,r'}$ 
              & $92.0_{\pm2.7}$ & $93.1_{\pm2.4}$ 
              & $91.2_{\pm2.7}$ & $93.7_{\pm2.3}$ 
              & $92.1_{\pm2.5}$ & $89.5_{\pm2.8}$ 
              & $91.8_{\pm1.5}$ & $91.9_{\pm1.5}$ \\[1ex]
      & 
          & RoBERTEye Fixations 
              & $E_S^{W,r},E_S^{W,r'}$ 
              & $85.5_{\pm3.3}$ & $90.5_{\pm2.8}$ 
              & $87.8_{\pm3.1}$ & $92.0_{\pm2.6}$ 
              & $88.0_{\pm2.8}$ & $88.3_{\pm2.9}$ 
              & $87.1_{\pm1.8}$ & $90.1_{\pm1.6}$ \\[1ex]
      & 
          & RoBERTEye Words 
              & $E_S^{W,r},E_S^{W,r'}$ 
              & $91.8_{\pm2.6}$ & $93.3_{\pm2.4}$ 
              & $91.0_{\pm2.7}$ & $93.7_{\pm2.3}$ 
              & $91.5_{\pm2.4}$ & $90.2_{\pm2.5}$ 
              & $91.5_{\pm1.6}$ & $92.3_{\pm1.5}$ \\
    \cmidrule(lr){2-12}
      & \multirow{8}{*}{Single Trial} 
          & \multirow{2}{*}{Reading Speed}  
              & $E_S^{W,r}$ 
              & $68.5_{\pm3.3}$ & $62.0_{\pm3.2}$ 
              & $69.2_{\pm3.3}$ & $60.8_{\pm3.2}$ 
              & $67.7_{\pm3.1}$ & $63.0_{\pm3.0}$ 
              & $68.4_{\pm1.9}$ & $62.0_{\pm1.7}$ \\[1ex]
      & 
          & 
              & $E_{EZ}^{W,1},E_S^{W,r}$ 
              & $69.9_{\pm3.2}$ & $65.9_{\pm2.7}$ 
              & $69.2_{\pm3.3}$ & $64.7_{\pm2.6}$ 
              & $67.4_{\pm3.0}$ & $64.6_{\pm2.5}$ 
              & $68.7_{\pm1.9}$ & $65.0_{\pm1.5}$ \\ \cmidrule{3-12}
      & 
          & \multirow{2}{*}{XGBoost}  
              & $E_S^{W,r}$ 
              & $72.1_{\pm3.1}$ & $70.5_{\pm2.5}$ 
              & $73.2_{\pm3.0}$ & $70.2_{\pm2.5}$ 
              & $70.8_{\pm2.8}$ & $69.0_{\pm2.3}$ 
              & $72.0_{\pm1.8}$ & $69.9_{\pm1.4}$ \\[1ex]
      & 
          & 
              & $E_{EZ}^{W,1},E_S^{W,r}$ 
              & $72.3_{\pm3.1}$ & $64.6_{\pm3.2}$ 
              & $72.5_{\pm3.3}$ & $67.4_{\pm3.1}$ 
              & $70.2_{\pm3.0}$ & $66.9_{\pm1.8}$ 
              & $71.5_{\pm1.8}$ & $71.1_{\pm2.2}$ \\ \cmidrule{3-12}
      & 
          & \multirow{2}{*}{RoBERTEye Fixations}   
              & $E_S^{W,r}$ 
              & $75.3_{\pm3.2}$ & $71.3_{\pm2.6}$ 
              & $76.3_{\pm3.1}$ & $71.1_{\pm2.6}$ 
              & $72.8_{\pm2.9}$ & $70.3_{\pm2.4}$ 
              & $74.7_{\pm1.7}$ & $70.9_{\pm1.4}$ \\[1ex]
      & 
          & 
              & $E_{EZ}^{W,1},E_S^{W,r}$ 
              & $73.2_{\pm3.1}$ & $72.5_{\pm2.4}$ 
              & $73.7_{\pm3.1}$ & $71.2_{\pm2.4}$ 
              & $71.6_{\pm2.9}$ & $71.0_{\pm2.3}$ 
              & $72.7_{\pm1.8}$ & $71.5_{\pm1.4}$ \\ \cmidrule{3-12}
      & 
          & \multirow{2}{*}{RoBERTEye Words} 
              & $E_S^{W,r}$ 
              & $76.3_{\pm3.1}$ & $71.2_{\pm2.6}$ 
              & $77.7_{\pm2.9}$ & $70.8_{\pm2.5}$ 
              & $75.2_{\pm2.8}$ & $69.5_{\pm2.4}$ 
              & $76.3_{\pm1.7}$ & $71.1_{\pm1.5}$ \\[1ex]
      & 
          &  
              & $E_{EZ}^{W,1},E_S^{W,r}$ 
              & $74.5_{\pm3.2}$ & $71.1_{\pm2.5}$ 
              & $75.9_{\pm3.2}$ & $71.7_{\pm2.5}$ 
              & $72.8_{\pm3.0}$ & $70.5_{\pm2.3}$ 
              & $74.3_{\pm1.7}$ & $71.1_{\pm1.5}$ \\
    \midrule
    \multirow{12}{*}{Test} 
      & \multirow{4}{*}{Paired Trials} 
          & Reading Speed 
              & $E_S^{W,r},E_S^{W,r'}$ 
              & $88.4_{\pm2.9}$ & $87.2_{\pm3.0}$ 
              & $88.3_{\pm2.9}$ & $87.6_{\pm3.0}$ 
              & $88.4_{\pm2.8}$ & $86.1_{\pm3.0}$ 
              & $88.3_{\pm1.7}$ & $86.9_{\pm1.8}$ \\[1ex]
      & 
          & XGBoost 
              & $E_S^{W,r},E_S^{W,r'}$ 
              & $90.5_{\pm2.6}$ & $93.0_{\pm2.3}$ 
              & $91.4_{\pm2.5}$ & $89.3_{\pm2.7}$ 
              & $91.9_{\pm2.4}$ & $91.6_{\pm1.5}$ 
              & $91.2_{\pm1.5}$ & $91.5_{\pm1.7}$ \\[1ex]
      & 
          & RoBERTEye Fixations 
              & $E_S^{W,r},E_S^{W,r'}$ 
              & $82.8_{\pm3.3}$ & $88.6_{\pm2.9}$ 
              & $83.7_{\pm3.1}$ & $88.8_{\pm2.9}$ 
              & $82.4_{\pm3.3}$ & $88.7_{\pm2.8}$ 
              & $82.9_{\pm1.9}$ & $88.7_{\pm1.6}$ \\[1ex]
      & 
          & RoBERTEye Words 
              & $E_S^{W,r},E_S^{W,r'}$ 
              & $87.6_{\pm3.0}$ & $89.6_{\pm2.7}$ 
              & $88.2_{\pm2.8}$ & $90.7_{\pm2.6}$ 
              & $89.6_{\pm2.6}$ & $87.9_{\pm2.9}$ 
              & $88.4_{\pm1.6}$ & $89.4_{\pm1.6}$ \\
    \cmidrule(lr){2-12}
      & \multirow{8}{*}{Single Trial} 
          & \multirow{2}{*}{Reading Speed} 
              & $E_S^{W,r}$ 
              & $69.2_{\pm2.9}$ & $60.9_{\pm3.1}$ 
              & $69.1_{\pm3.0}$ & $61.9_{\pm3.1}$ 
              & $68.4_{\pm3.1}$ & $62.4_{\pm3.2}$ 
              & $68.9_{\pm1.8}$ & $61.7_{\pm1.7}$ \\[1ex]
      & 
          & 
              & $E_{EZ}^{W,1},E_S^{W,r}$ 
              & $69.4_{\pm3.2}$ & $61.8_{\pm3.1}$ 
              & $69.4_{\pm3.1}$ & $62.2_{\pm2.9}$ 
              & $68.0_{\pm3.1}$ & $62.6_{\pm3.1}$ 
              & $68.9_{\pm1.8}$ & $62.2_{\pm1.8}$ \\ \cmidrule{3-12}
      & 
          & \multirow{2}{*}{XGBoost} 
              & $E_S^{W,r}$ 
              & $70.7_{\pm2.9}$ & $66.7_{\pm3.0}$ 
              & $72.2_{\pm2.9}$ & $67.6_{\pm3.0}$ 
              & $69.7_{\pm2.9}$ & $66.0_{\pm3.1}$ 
              & $70.8_{\pm1.7}$ & $66.7_{\pm1.2}$ \\[1ex]
      & 
          & 
              & $E_{EZ}^{W,1},E_S^{W,r}$ 
              & $72.2_{\pm3.0}$ & $65.5_{\pm2.9}$ 
              & $72.7_{\pm2.9}$ & $68.0_{\pm2.8}$ 
              & $70.1_{\pm2.9}$ & $67.2_{\pm3.0}$ 
              & $71.6_{\pm1.7}$ & $70.1_{\pm2.1}$ \\ \cmidrule{3-12}
      & 
          & \multirow{2}{*}{RoBERTEye Fixations}
              & $E_S^{W,r}$ 
              & $72.1_{\pm2.9}$ & $64.8_{\pm3.0}$ 
              & $72.9_{\pm3.0}$ & $64.1_{\pm3.0}$ 
              & $71.5_{\pm3.0}$ & $65.8_{\pm3.0}$ 
              & $72.1_{\pm1.7}$ & $64.8_{\pm1.7}$ \\[1ex]
      & 
          & 
              & $E_{EZ}^{W,1},E_S^{W,r}$ 
              & $70.5_{\pm3.0}$ & $67.7_{\pm2.8}$ 
              & $70.7_{\pm2.9}$ & $65.9_{\pm3.0}$ 
              & $68.6_{\pm3.0}$ & $68.0_{\pm2.9}$ 
              & $69.8_{\pm1.7}$ & $67.2_{\pm1.7}$ \\ \cmidrule{3-12}
      & 
          & \multirow{2}{*}{RoBERTEye Words} 
              & $E_S^{W,r}$ 
              & $71.9_{\pm3.0}$ & $64.6_{\pm3.0}$ 
              & $74.0_{\pm3.0}$ & $64.8_{\pm2.9}$ 
              & $71.2_{\pm3.1}$ & $63.8_{\pm2.9}$ 
              & $72.3_{\pm1.7}$ & $64.3_{\pm1.8}$ \\[1ex]
      & 
          & 
              & $E_{EZ}^{W,1},E_S^{W,r}$ 
              & $73.8_{\pm3.1}$ & $63.0_{\pm3.0}$ 
              & $73.0_{\pm3.0}$ & $62.5_{\pm3.0}$ 
              & $72.4_{\pm3.0}$ & $63.5_{\pm3.0}$ 
              & $73.0_{\pm1.8}$ & $63.1_{\pm1.8}$ \\
    \bottomrule
  \end{tabular}%
}
\caption{Precision and Recall (repeated reading being positive and first reading being negative) results for the two variants of the first vs. second reading prediction task with 95\% confidence intervals, aggregated across 10 cross-validation splits, and presented for both test and validation partitions. $E_{EZ}^{W,1}$ denotes synthesized eye movements generated using \mbox{E-Z} Reader \cite{reichle2003ez}.}
\label{app:tab:prec_recall}
\end{table*}

\end{document}